%% file: TMM.tex
\documentclass[lettersize,journal]{IEEEtran}
\usepackage{amsmath,amsfonts}
\usepackage{algorithmic}
\usepackage{algorithm}
\usepackage{array}
\usepackage[caption=false,font=normalsize,labelfont=sf,textfont=sf]{subfig}
\usepackage{textcomp}
\usepackage{stfloats}
\usepackage{url}
\usepackage{verbatim}
\usepackage{graphicx}
\usepackage{cite}
\usepackage{booktabs}
\usepackage{multirow}
\usepackage{arydshln}
\usepackage{amssymb}
\usepackage{hyperref}
\hypersetup{
    colorlinks=true, 
    linkcolor=black,
    urlcolor=blue,
    citecolor=black,
}

\begin{document}

\title{Generalizable and Adaptive Continual Learning Framework for AI-generated Image Detection}

\author{Hanyi Wang, Jun Lan, Yaoyu Kang, Huijia Zhu, Weiqiang Wang, Zhuosheng Zhang, \\ $^\dagger$Shilin Wang,~\IEEEmembership{Senior Member,~IEEE,}
\thanks{H. Wang, Y. Kang, Z. Zhang and S. Wang are affiliated with the School of Electronic Information and Electrical Engineering, Shanghai Jiao Tong University, Shanghai 200240, China (e-mail: why$\_$820@sjtu.edu.cn; kangyaoyu@sjtu.edu.cn; zhangzs@sjtu.edu.cn; wsl@sjtu.edu.cn).}
\thanks{J. Lan, H. Zhu, and W. Wang are with Antgroup (e-mail: yelan.lj@antgroup.com; huijia.zhj@antfin.com; wang.weiqiang@gmail.com).}
\thanks{$^\dagger$Shilin Wang is the corresponding author.}
}

\markboth{Journal of \LaTeX\ Class Files,~Vol.~14, No.~8, August~2021}%
{Shell \MakeLowercase{\textit{et al.}}: A Sample Article Using IEEEtran.cls for IEEE Journals}

\maketitle

\begin{abstract}
The malicious misuse and widespread dissemination of AI-generated images pose a significant threat to the authenticity of online information. Current detection methods often struggle to generalize to unseen generative models, and the rapid evolution of generative techniques continuously exacerbates this challenge. Without adaptability, detection models risk becoming ineffective in real-world applications. To address this critical issue, we propose a novel three-stage domain continual learning framework designed for continuous adaptation to evolving generative models. In the first stage, we employ a strategic parameter-efficient fine-tuning approach to develop a transferable offline detection model with strong generalization capabilities. Building upon this foundation, the second stage integrates unseen data streams into a continual learning process. To efficiently learn from limited samples of novel generated models and mitigate overfitting, we design a data augmentation chain with progressively increasing complexity. Furthermore, we leverage the Kronecker-Factored Approximate Curvature (K-FAC) method to approximate the Hessian and alleviate catastrophic forgetting. Finally, the third stage utilizes a linear interpolation strategy based on Linear Mode Connectivity, effectively capturing commonalities across diverse generative models and further enhancing overall performance. We establish a comprehensive benchmark of 27 generative models, including GANs, deepfakes, and diffusion models, chronologically structured up to August 2024 to simulate real-world scenarios. Extensive experiments demonstrate that our initial offline detectors surpass the leading baseline by +5.51\% in terms of mean average precision. Our continual learning strategy achieves an average accuracy of 92.20\%, outperforming state-of-the-art methods. 
\end{abstract}


\input{Section/1_intro}
\input{Section/2_related_work}
\input{Section/3_analysis}
\input{Section/4_method}
\input{Section/5_experiments}

\input{Section/6_conclusion}
\input{Section/7_acknowledgement}

\bibliographystyle{IEEEtran}
\bibliography{main}

\vspace{-1cm}
\input{Section/8_authors}
\end{document}

%% file: Section/1_intro.tex
\section{Introduction}
\IEEEPARstart{R}{ecent} advancements in diffusion models\cite{ho2020denoising, midjourney, Murahari2021DALLECI, rombach2022high} and natural language processing \cite{touvron2023llama, devlin2018bert} have significantly propelled the field of image synthesis. While these technological innovations offer substantial benefits, they also introduce serious risks, notably the potential for spreading disinformation and creating fraudulent identities. Given these critical concerns, it is imperative to develop practical, robust techniques for the detection of AI-generated images.

To address this need, we first analyze the specific challenges in developing a practical AI-generated image detector: (1) \textbf{Generalization to Unseen Generative Models.}  Existing detection methods for AI-generated images often target specific artifacts of known generative techniques. For instance, Wang \textit{et al.}\cite{wang2020cnn} and Tan \textit{et al.}\cite{tan2023learning} focused on detecting GAN-generated images, while Wang \textit{et al.}\cite{wang2023dire} and Luo \textit{et al.}\cite{luo2024lare} concentrated on identifying images from diffusion models. However, these tailored methods often struggle to generalize to images generated by new or significantly altered techniques. Recent research\cite{ojha2023towards, tan2024rethinking, li2024improving, liu2024forgery} has explored cross-generalization by investigating universal artifacts and employing sophisticated feature extractors. Despite these advances, such methods remain susceptible to overfitting due to their over-reliance on specific generative patterns. Consequently, there is a critical need for detection methods with robust underlying feature representations capable of identifying common forgery artifacts across diverse generative models. (2) \textbf{Adaptation to Evolving Generative Models.} The real-world application scenario of generative techniques is highly dynamic and rapidly evolving, with new models being introduced regularly. This continuous emergence presents ongoing challenges in maintaining the generalization capabilities of detection systems, especially when the characteristic distribution of data generated by new models diverges significantly from the initial training phase. Without sufficient adaptability to these evolving generative techniques, existing detection models risk becoming progressively ineffective in real-world applications. A straightforward approach to address this issue is to train a new model from scratch, incorporating both new and original data. While this method may seem feasible for a small number of training instances, it becomes increasingly inefficient and resource-intensive as the dataset expands and new generative models proliferate. Consequently, establishing an effective and efficient mechanism for model updates is crucial to ensure the sustained effectiveness of detection systems.

To address these challenges, we propose a three-stage domain continual learning framework specifically designed for detecting AI-generated images. This framework comprises a transferable offline detection model with high initial generalization capability, which is then dynamically updated to keep pace with evolving generative models.

We begin by proposing a simple yet effective detector with strong initial generalization capabilities. Recent work \cite{ojha2023towards} attributed the decline in generalization performance to the development of asymmetrical decision boundaries in traditional models. These skewed boundaries bias models towards misclassifying unseen forgeries from outside the training distribution as authentic. The key insight to mitigate this issue is to perform classification within a feature space unbiased towards distinguishing images based on these skewed classes. This strategy aims to neutralize feature biases, promoting more balanced detection across diverse forgery types. Building upon this perspective, our work extends this line of research. However, instead of employing a model with frozen non-linearities using linear probing, as in Ojha \textit{et al.}\cite{ojha2023towards},  we introduce subtle adjustments to enable the effective learning of forgery-specific features while preserving the unbiased feature space crucial for robust generalization.

To achieve this, we propose a strategic parameter-efficient fine-tuning approach. Rather than employing adapters that introduce forgery-aware priors \cite{luo2023forgery, shao2023deepfake, liu2024forgery}, which can lead to subjective biases, we employ Low-Rank Adaptation (LoRA) \cite{hu2022lora}, where LoRA is valued for its capacity to adapt models directly and efficiently without relying on prior assumptions. Specifically, we focus on fine-tuning the Multi-Layer Perceptron (MLP) layers within the Transformer blocks of the CLIP Vision Transformer (CLIP-ViT) model. These layers are pivotal because they function similarly to key-value memories in Transformer-based models \cite{geva2020transformer}, effectively integrating and storing task-specific knowledge within ViT. This targeted fine-tuning strategy significantly boosts the initial detection performance to generalize across various generative models.

Subsequently, we propose a novel continual learning method specifically designed for AI-generated image detection, addressing two critical domain-specific challenges. First, there is an inherent difficulty in collecting sufficient representative samples from genuinely novel generative models, as only a limited number of examples are often available, and many of these models are not open-sourced. This scarcity of data poses significant challenges to the effective learning of new knowledge. Second, while images generated by different methods (e.g., GANs, diffusion models, autoencoders) exhibit distinct feature distributions, they fundamentally share the characteristic of being synthetic. Effectively leveraging these shared characteristics across diverse feature distributions is essential for achieving robust generalization. 

To address these challenges, our method comprises two key innovations. To mitigate data scarcity, we design a data augmentation chain with progressively increasing complexity to enrich data diversity, enabling efficient learning from limited samples and reducing the risk of overfitting. To leverage the shared synthetic features of AI-generated images, we introduce Linear Mode Connectivity (LMC) within the loss landscape. This involves modeling relationships between new and old detection models by exploring low-loss pathways during the continual learning process. Our empirical analysis confirms the presence of Linear Mode Connectivity in continual learning for AI-generated image detection. Subsequently, we extend our study with a theoretical analysis to establish these pathways. Based on the theoretical foundation, we propose a straightforward linear interpolation strategy using LMC to capture essential commonalities across diverse generative models, thereby enhancing generalization and adaptability. Furthermore, our theoretical investigation of these LMC-based pathways reveals that constraining the Hessian matrix plays a crucial role in mitigating catastrophic forgetting. Consequently, we employ Kronecker-Factored Approximate Curvature (K-FAC)\cite{martens2015optimizing} for Hessian approximation to provide a more rigorous theoretical grounding for these pathways.

Moreover, we establish a comprehensive benchmark that emulates an online setting. This benchmark consists of a dataset comprising 27 well-known generative models, including GANs, deepfakes, and diffusion models, with release dates spanning from October 2017 up to August 2024. The chronological ordering of these models ensures that the benchmark accurately reflects the evolution of generative models in the real world. Our contributions are summarized as follows:

\begin{itemize}
\item{We propose a simple yet effective AI-generated image detector with high generalization capabilities across diverse generative models.}
\item{We introduce a novel, model-agnostic continual learning framework for AI-generated image detection that can effectively update models based on only a few new samples.}
\item{We develop a comprehensive benchmark for continual AI-generated image detection, including public datasets featuring GANs, deepfakes, and diffusion models.}
\item{We conduct extensive experiments on our benchmark, demonstrating the superior generalization of our offline detectors and the enhanced continual learning performance compared to state-of-the-art methods.}
\end{itemize}

%% file: Section/2_related_work.tex
\section{Related Work}
\subsection{AI-generated Image Detection}
A significant challenge in AI-generated image detection methods is their generalization ability to unseen breeds of generative models. Wang $et$ $al.$\cite{wang2020cnn} established an effective baseline in detecting images generated by various GAN models. Ojha $et$ $al.$ \cite{ojha2023towards} employed nearest neighbor and linear probing classifiers on pre-trained CLIP \cite{radford2021learning} visual features, achieving promising results across different generative families. More recently, Tan $et$ $al.$ \cite{tan2024rethinking} introduced a novel artifact representation focusing on local up-sampling artifacts in image pixels. Liu $et$ $al.$ \cite{liu2024forgery} designed a forgery-aware adaptive transformer that refines image features to detect forgery traces across both image and frequency domains. Li $et$ $al.$ \cite{li2024improving} proposed a lightweight detector employing three simple image transformations. Despite these advances, these methods demonstrate limited generalization, particularly when confronted with unseen generative architectures or rapidly evolving synthesis techniques.

\subsection{Continual Learning}
Continual learning methods can be broadly categorized into three groups: rehearsal-based \cite{chaudhry2019tiny}, architecture-based \cite{mallya2018packnet}, and regularization-based \cite{kirkpatrick2017overcoming} approaches. While the application of continual learning to AI-generated image detection is still nascent, it is critically important and urgently needed in real-world scenarios. Recent studies \cite{pan2023dfil, sun2023continual} have explored continual learning for face forgery detection, and Li \textit{et al.} \cite{li2023continual} introduced a continual deepfake detection benchmark encompassing GANs and deepfakes. For diffusion models, Epstein $et$ $al.$\cite{epstein2023online} suggested an online setting, though it requires extensive resources to train with all accumulated historical images. In this paper, we propose a novel model updating mechanism specifically for the continual detection of AI-generated images. Furthermore, we establish a comprehensive benchmark that includes GANs, deepfakes and diffusion models, structured chronologically to accurately reflect real-world scenarios.

\subsection{Linear Mode Connectivity}
Linear mode connectivity is a phenomenon where different minima in a high-dimensional, non-convex objective landscape can be connected by low-loss paths in the parameter space. Frankle $et$ $al.$\cite{frankle2020linear} showed that, when starting from the same initialization, different minima of the same task can be connected by a low-pass path. Building upon this discovery, subsequent research \cite{mirzadeh2020linear} demonstrated that solutions for multitask and continual learning can be connected by a linear path with low errors, termed Linear Mode Connectivity. This suggests a promising strategy for adapting models to new tasks while retaining previously learned knowledge. In this paper, we explore the application of linear mode connectivity to address the challenge of discerning and leveraging commonalities across different generative models. By modeling the relationships between new and old detection models and identifying low-loss pathways during the continual learning process, we effectively uncover and utilize these common characteristics.

%% file: Section/3_analysis.tex
\section{Continual Learning Analysis for AI-Generated Image Detection}
\subsection{Problem Formulation}In this section, we first present a basic formulation for the continual learning of AI-generated image detection. Let a sequential series of detection tasks be defined as $D = \{D_1, D_2, \dots, D_T\}$, where each task, $D_t$, comprises a set of input-label pairs $\{(x_i, y_i)\}_{i=1}^{N_t}$. Here, $x_i$ denotes the input image, $y_i$ represents the binary label indicating whether the image is authentic or AI-generated, and $N_t$ is the total number of samples in task $t$. At each step $t$, both the dataset $D_t$ for the current task and the previous model $\theta_{1:(t-1)}$ are provided. The primary challenge is to update the previous model $\theta_{1:(t-1)}$ to $\theta_{1:t}$, focusing on two critical objectives: 1) plasticity: the capacity of the model to acquire new knowledge from $D_t$; and 2) stability: the ability to preserve previously learned knowledge from earlier tasks $D_{1:(t-1)}$.

\begin{figure}[t!]
  \centering
  \setlength{\abovecaptionskip}{-0.02cm}
  \includegraphics[width=\linewidth]{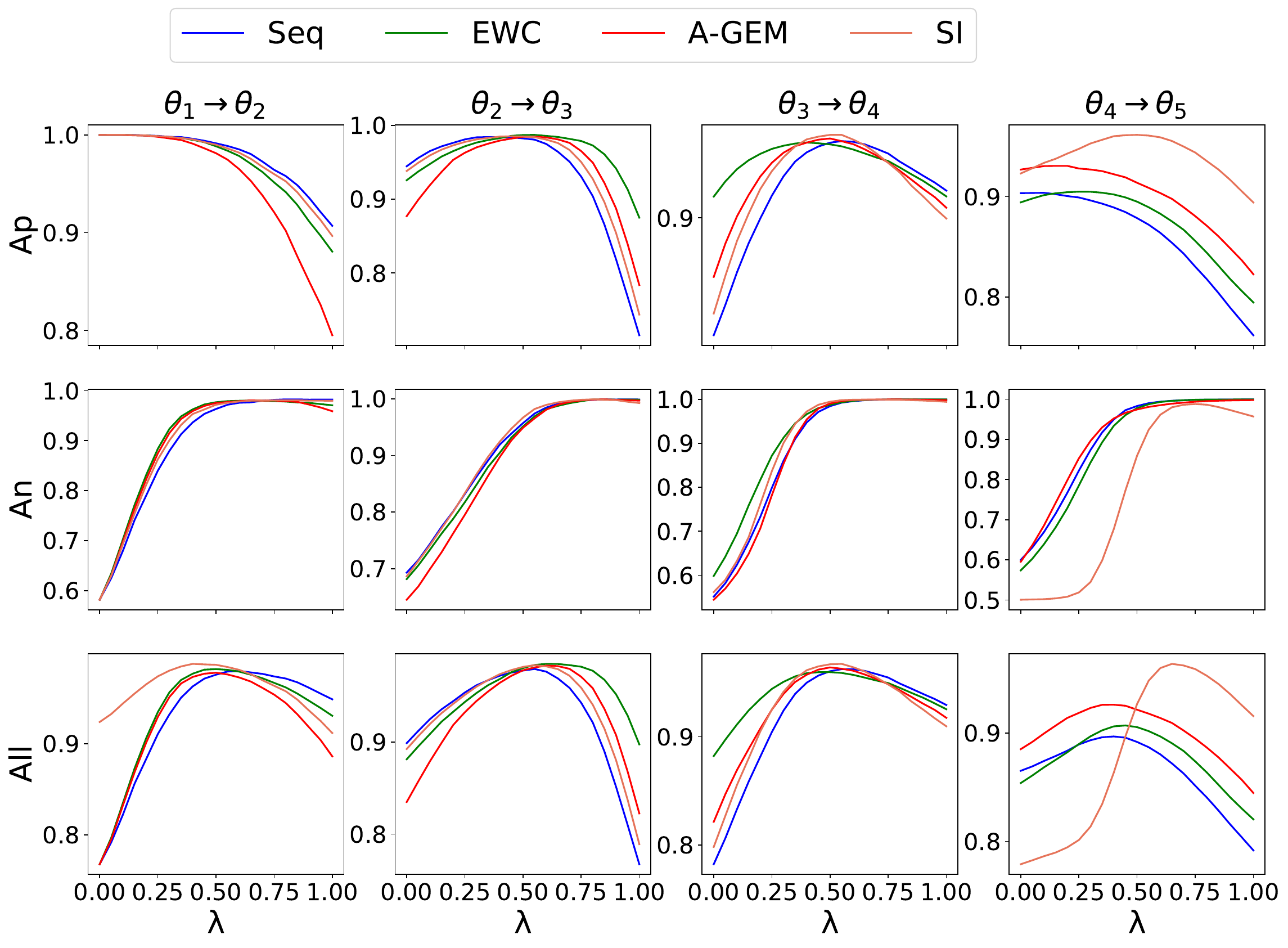}
  \caption{\textbf{Testing Accuracy Curves.} Accuracy curves for four continual learning baselines (Seq, EWC\cite{kirkpatrick2017overcoming}, A-GEM\cite{chaudhry2018efficient}, and SI\cite{zenke2017continual}) showing the effects of linear interpolation between two adjacent minima on historical tasks (Ap), current tasks (An), and a combined dataset of all tasks (All). The parameter $\lambda$ represents the interpolation factor.}
  \label{fig:fig_1}
\end{figure}

\subsection{Mode Connectivity Evaluation}

In this section, we explore mode connectivity in the context of continual learning for AI-generated image detection. Given the minima of two consecutive tasks, denoted as $\theta_{1:(t-1)}$ and $\theta_{t}$, where $\theta_{1:(t-1)}$ servers as the initial parameters for task $t$. We hypothesize that there exists a linear path in the parameter space effectively connecting these minima without significant loss increases, formulated as:
\begin{equation}
  \phi(\lambda) = (1-\lambda) \cdot \theta_{1:(t-1)} + \lambda \cdot \theta_{t}, 
  \quad \lambda \in [0, 1].
\end{equation}

\begin{figure*}[t!]
  \centering
  \includegraphics[width=0.9\textwidth]{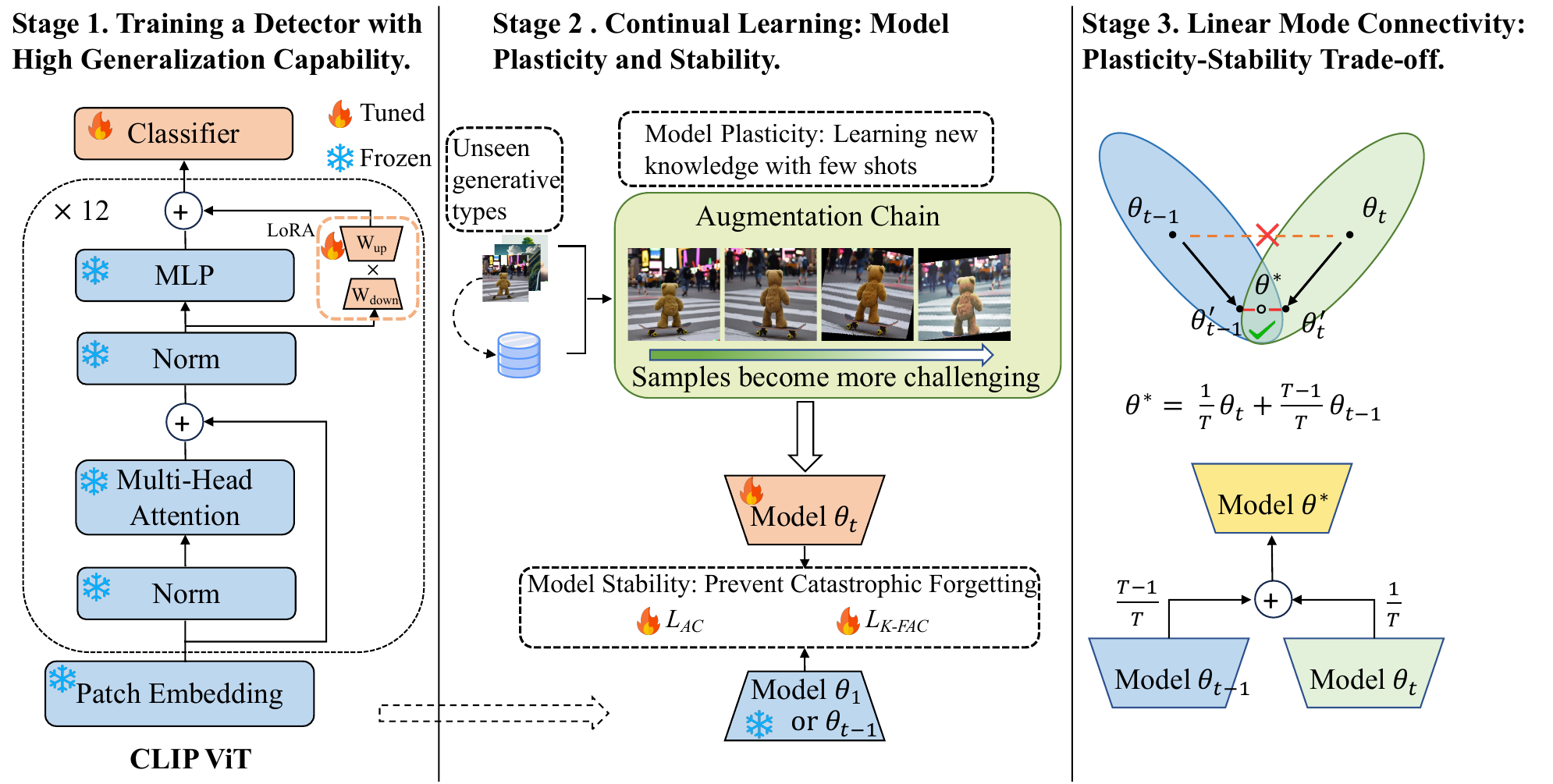}
  \caption{\textbf{Framework Overview.} A three-stage continual learning framework for AI-generated image detection is presented. Specifically, the first stage employs parameter-efficient fine-tuning strategies using LoRA on MLP layers to develop a generalizable offline detector. The second stage integrates unseen data streams into the continual learning process, proposing a data augmentation chain and leveraging the K-FAC method to acquire new knowledge and prevent catastrophic forgetting. The final stage utilizes linear interpolation to discern and exploit commonalities across different generative models, aiming to balance mode plasticity and stability.}
  \label{fig:fig_2}
\end{figure*}

Here, $\lambda$ is a factor that linearly interpolates between the parameters of two adjacent tasks. By traversing this linear path, $\phi(\lambda)$, we evaluate the interpolation performance between $\theta_{1:(t-1)}$ and $\theta_{t}$. For space considerations, we report the performance of the first four tasks, with detailed descriptions of the task sequence and experimental setups reserved for the experiment section. Figure \ref{fig:fig_1} illustrates testing accuracy curves along the linear path between two adjacent minima for historical tasks, current tasks, and all tasks, denoted as Ap, An, and All, respectively. Our observations indicate that: (1) Performance on previous tasks (Ap) improves along the trajectory from $\theta_{1:(t-1)}$ to $\theta_{t}$ compared to the initial point, suggesting that parameter adjustments along this path can enhance memorization. (2) There are intervals along the path where performance on the current task (An) matches that observed at the endpoints. (3) There are points along the path yielding superior overall performance, indicating that a better trade-off between stability and plasticity can be achieved along this linear interpolation than at the endpoints.

Therefore, by pulling the path from the high-loss ridge to the low-loss region, we can achieve higher accuracy than at the endpoints. In the next subsection, we will present a theoretical analysis to demonstrate how to construct a high-accuracy pathway for all tasks.

\subsection{Theoretical Analysis}
To construct a high-accuracy pathway for all tasks, we first quantify the phenomenon of forgetting across two sequential tasks. Let $\theta_t$ be the optimal weight for task $t$, trained exclusively on that task, and let $L_t(\theta)$ denote the empirical loss for task $t$. We define forgetting from the end of task $D_{t-1}$ to the end of task $D_t$ as:
\begin{equation}
  F_{t-1} \triangleq L_{t-1}(\theta_t) - L_{t-1}(\theta_{t-1}).
\end{equation}

Assuming a second-order Taylor expansion of the loss function around the convergence points of previous tasks, we estimate $F_{t-1}$ by approximating $L_{t-1}(\theta_t)$ around $\theta_{t-1}$:
\begin{align}
  L_{t-1}(\theta_t) & \approx  L_{t-1}(\theta_{t-1}) + (\theta_{t}-\theta_{t-1})^{T} \nabla L_{t-1}(\theta_{t-1}) \notag \\ 
  & + \frac{1}{2}(\theta_{t}-\theta_{t-1})^{T} \nabla^{2} L_{t-1}(\theta_{t-1}) (\theta_{t}-\theta_{t-1}) \notag \\
  & \approx L_{t-1}(\theta_{t-1}) \notag \\ 
  & + \frac{1}{2}(\theta_{t}-\theta_{t-1})^{T} \nabla^{2} L_{t-1}(\theta_{t-1}) (\theta_{t}-\theta_{t-1}),
\end{align}

\noindent
where $\nabla L_{t-1}(\theta_{t-1}) \approx 0$ because $\theta_{t-1}$ is at a local minimum and $\nabla^{2} L_{t-1}(\theta_{t-1})$ is the Hessian matrix of the loss function at $\theta_{t-1}$. Thus, the forgetting $F_{t-1}$ can be approximated as:
\begin{align}
    F_{t-1} & = L_{t-1}(\theta_t) - L_{t-1}(\theta_{t-1}) \notag \\ 
    & \approx \frac{1}{2}(\theta_{t}-\theta_{t-1})^{T} \nabla^{2} L_{t-1}(\theta_{t-1}) (\theta_{t}-\theta_{t-1}).
\end{align}

This indicates that the extent of forgetting is influenced by the Hessian's curvature and the magnitude of the weight changes between successive models. Minimizing both the displacement and curvature is crucial to reduce forgetting.

We further establish a theoretical upper bound for the cumulative empirical loss across all tasks by considering multitask learning as an oracle benchmark:
\begin{equation}
  \theta^* = \underset{\theta}{\arg\min} \sum_{t=1}^T L_t(\theta).
\end{equation}

For each task, we estimate the loss $L_{t}(\theta^*)$ near $\theta_{t}$:
\begin{align}
  L_t(\theta^*) \approx & L_t(\theta_t) + \frac{1}{2}(\theta^*-\theta_t)^{T} \nabla^{2} L_t(\theta_t)(\theta^*-\theta_t) \notag \\
  \leq & L_t(\theta_t) + \frac{1}{2} \lambda_t^{max} \Vert \theta^*-\theta_t \Vert^2,
\end{align} 
\noindent
where$\lambda_{t}^{max}$ is the maximum eigenvalue of the Hessian matrix for task $t$. Aggregating these inequalities across all tasks, we obtain an upper bound for the total empirical loss:

\begin{align}
\label{e_5}
\sum_{t=1}^{T} L_t(\theta^*) \leq & \sum_{t=1}^{T} L_t(\theta_t) + \frac{1}{2} \sum_{t=1}^{T}\lambda_t^{max} \left\| \theta^* - \theta_t \right\|^2 \\
\label{e_6}
\leq & \sum_{t=1}^{T} L_t(\theta_t) + \frac{1}{2} \lambda^{max}\sum_{t=1}^{T} \left\| \theta^* - \theta_t \right\|^2,
\end{align}
\noindent 
where $\lambda^{max}=max(\lambda_1^{max},\lambda_2^{max},\dots,\lambda_T^{max})$. Thus, the optimal weight for continual learning can be expressed as:
\begin{equation}
  \theta^* = \frac{1}{T} \sum_{t=1}^{T} \theta_t.
\end{equation}

This formulation suggests that the optimal weight can be derived by linear interpolation, consistent with the principles of linear mode connectivity. Moreover, considering $\theta_{t-1}$ and $\theta_{t}$ as optimal solutions for the previous $t-1$ tasks and the current $t$-th task respectively, we further deduce:
\begin{equation}
\theta^* = \frac{T-1}{T} \theta_{t-1} + \frac{1}{T} \theta_{t}.
\end{equation}

%% file: Section/4_method.tex
\section{Method}

Based on the preceding analysis, we propose a three-stage continual learning framework for AI-generated image detection, as illustrated in Figure \ref{fig:fig_2}. In the first stage, we employ a strategic parameter-efficient tuning approach to develop a robust offline detector with strong generalization capabilities across various generative techniques. Subsequently, the second stage integrates unseen data streams into a continual learning process. To efficiently learn new knowledge from limited samples of novel generative models, we design a data augmentation chain with progressive complexity. Furthermore, to mitigate catastrophic forgetting, we leverage the K-FAC\cite{martens2015optimizing} method for Hessian approximation, thereby reducing the risk of forgetting previously learned knowledge. Finally, the third stage leverages linear interpolation based on linear mode connectivity to discern and utilize commonalities across different generative models, achieving a better trade-off between model plasticity and stability.

\subsection{Stage 1. Training a Detector with High Generalization Capability} 

This stage aims to establish a foundational offline detection model with strong generalization capabilities. Our objective is to enable effective learning of forgery-specific features while preserving the pre-trained model's robust generalization. To achieve this, we employ parameter-efficient fine-tuning using LoRA\cite{hu2022lora}. Unlike methods employing adapters that introduce forgery-aware priors \cite{luo2023forgery, shao2023deepfake, liu2024forgery}, which can lead to subjective biases, LoRA adapts models directly and efficiently without relying on prior assumptions. Specifically, we target the MLP layers within the Transformer blocks of a CLIP-ViT model. These layers are pivotal, functioning similarly to key-value memories in Transformer-based models \cite{geva2020transformer} and effectively integrating and storing task-specific knowledge within ViT. This targeted approach allows for subtle yet effective adjustments, enhancing the generalization capabilities to detect a broader range of AI-generated images.

\subsection{Stage 2. Continual Learning: Model Plasticity and Stability}

When existing offline detection models struggle to identify new generative content, updating them with new data streams becomes essential. This adaptation requires a careful balance between model plasticity and stability.

\subsubsection{Model Plasticity}

To effectively acquire knowledge from a limited number of newly generated samples, we propose the Augmentation Chain, a strategic approach explicitly designed to enrich training data diversity. This strategy employs data augmentation techniques to create a sequence of samples with progressively increasing complexity, thereby gradually introducing more intricate challenges. This transition from simpler to more complex samples not only improves the model's adaptability to new tasks but also enhances its robustness against overfitting.

Specifically, our augmentation chain begins with basic geometric distortions, such as RandomFlip and RandomAffine. We then increase the transformation complexity by incorporating RandAugment, controlled by two hyperparameters, $N$ and $M$, to adjust the intensity and diversity of the augmentations. This chain includes three distinct augmentation steps, and we integrate these steps into the training process using the loss function:
\begin{equation}
    \mathcal{L}_{AC}(\theta_t)=\mathcal{L}_{cls}(\theta_t(x), y) + \lambda \sum_{i=1}^{3} \mathcal{L}_{cls}(\theta_t(x_i), y),
\end{equation}

\noindent
where $x_i$ represents the augmented input sample $x$ after $i$ augmentation steps, and $L_{cls}$ denotes the binary cross-entropy loss. $\lambda$ is a tuning parameter balancing the influence of augmented samples on the overall loss, ensuring the model learns robustly from both original and transformed data.

\subsubsection{Model Stability}

To minimize forgetting, our theoretical analysis emphasizes the importance of reducing both weight displacement and the curvature of the loss function. We propose using the K-FAC\cite{martens2015optimizing} method to estimate the Hessian matrix at each layer of a neural network, which provides a more accurate approximation while managing memory usage effectively.

Formally, for an $M$-layer neural network, the input $a_0=x$ progresses through the network to produce an output $h_M$. Each $m$-th layer transforms its input as $h_m=W_m a_{m-1}$, where $h_m$ is the pre-activation and $a_m = f_m(h_m)$ is the activation function applied elementwise. Using the chain rule, the Hessian $H^{(m)}$ concerning the weights at any layer $m$ can be factored into two small matrices, $\mathcal Q_m$ and $\mathcal H_m$, such that:
\begin{equation}
H^{(m)} = \frac{\partial^2 L}{\partial \operatorname{vec}(W_m) \partial \operatorname{vec}(W_m)^T} = \mathcal{Q}_m \otimes \mathcal{H}_m,
\end{equation}

\noindent
where $\mathcal{Q}_m = a_{m-1} a_{m-1}^T$ and $\mathcal{H}_m = \frac{\partial^2 L}{\partial h_m \partial h_m}$. Here, $L$ denotes the loss function, and vec$(W_m)$ is the vectorized form of the weight matrix at layer $m$. This decomposition allows for efficient computation, utilizing the Kronecker product to approximate the Hessian-vector product for each layer as:
\begin{equation}
\mathcal{Q}_m \otimes \mathcal{H}_m \operatorname{vec}(W_m - W_m^{\text{pre}}) = \operatorname{vec}(\mathcal{H}_m (W_m - W_m^{\text{pre}})^T \mathcal{Q}_m),
\end{equation}
\noindent
where $W_m^{\text{pre}}$ is the weight matrix before the update.

We then define the K-FAC loss for training adjustments:
\begin{equation}
\mathcal{L}_{\text{K-FAC}} = \frac{1}{2} \sum_{m=1}^M (\Delta W_m)^T (\mathcal{K}_m)^{-1} (\Delta W_m),
\end{equation}
where $\Delta W_m = W_m - W_{m}^{\text{pre}}$ and $\mathcal{K}_m = \mathcal{Q}_m \otimes \mathcal{H}_m$ represents the K-FAC approximation of the inverse Hessian for layer $m$.

The overall loss function combining the augmentation chain and K-FAC is then expressed as:
\begin{equation}
\mathcal{L}(\theta_t) = \mathcal{L}_{AC}(\theta_t) + \gamma \mathcal{L}_{\text{K-FAC}},
\end{equation}
where $\gamma$ is a regularization parameter balancing the trade-off between acquiring new knowledge and retaining previous learning. This approach mitigates catastrophic forgetting and stabilizes updates, preserving previously learned knowledge while accommodating new information.

\subsection{Stage 3. Linear Mode Connectivity: Plasticity-Stability Trade-off}

We now have two neural networks: $\theta_{1:(t-1)}$, which preserves previous knowledge, and $\theta_{t}$, the optimal weight for the current task. To both preserve crucial commonalities essential for effective generalization across diverse generative models and learn the unique forgery traces of new models, we model the relationship between these networks based on Linear Mode Connectivity. This approach identifies low-loss pathways during continual learning, allowing us to discern and leverage commonalities across different generative models, thus achieving a balance between plasticity and stability. We seamlessly integrate the strengths of both models using linear interpolation, as described by the following equation:
\begin{equation}
  \theta_{1:t} = \frac{T-1}{T} \theta_{1:(t-1)} + \frac{1}{T} \theta_{t}.
\end{equation}

%% file: Section/5_experiments.tex
\input{Table/table_1}
\section{Experiments}
\subsection{Experimental Setups}
\noindent
\textbf{Benchmark.} We established a comprehensive benchmark encompassing various families of generative models, including GANs, deepfakes, and diffusion models. Our dataset includes images sourced from 27 different models released between Oct. 2017 and Aug. 2024, as detailed in Table \ref{tab:dataset-table}. We meticulously organized these datasets to align with the chronological emergence of generative models, thus ensuring that our approach accurately reflects real-world scenarios in the continual learning context. For datasets that do not provide a predefined split, we divided the data into training and testing sets in a 1:9 ratio for continual learning tasks. For datasets that came with their own split, we preserved the original test sets and created our continual training sets by randomly sampling one-tenth of the test set size from the existing training sets.

Specifically, we collected GAN datasets from \cite{wang2020cnn}, including ProGAN\cite{karras2018progressive}, CycleGAN\cite{zhu2017unpaired}, StarGAN\cite{choi2018stargan}, BigGAN\cite{brock2019large}, StyleGAN\cite{karras2019style}, GauGAN\cite{park2019semantic}, and StyleGAN2\cite{karras2020analyzing}. For diffusion models, we sourced datasets from \cite{ricker2022towards, ojha2023towards, zhu2024genimage, chen2024drct, Li2024ImprovingSI}, featuring DDPM\cite{ho2020denoising}, ADM\cite{dhariwal2021diffusion}, iDDPM\cite{nichol2021improved}, DALL-E\cite{Murahari2021DALLECI}, GLIDE\cite{nichol2021glide}, LDM\cite{rombach2022high}, PNDM\cite{liu2022pseudo}, Wukong\cite{gu2022wukong}, SD v1.4/1.5\cite{rombach2022high}, VQDM\cite{gu2022vector}, SD v2.1\cite{rombach2022high}, SDXL v1.0\cite{podell2024sdxl}, SD-Turbo\cite{sauer2024fast}, SDXL-Turbo\cite{sauer2025adversarial}, SD v3.0\cite{esser2024scaling}, PixArt-$\Sigma$\cite{chen2024pixart} and FLUX.1\cite{FLUX}. Additionally, we included deepfake datasets, also obtained from \cite{wang2020cnn}. 

\input{Table/table_2}
\input{Table/table_3}

\noindent
\textbf{Implementation Details.} We employed two variants of CLIP-ViT (\textit{i.e.} ViT-B/32, ViT-L/14) as backbones to train the initial offline detector. For the LoRA hyperparameters, the rank $r$ was set to 8. During the continual learning phase, we utilized CLIP ViT-B/32. We monitored the model's performance on newly generated images and initiated the continual learning process by integrating previously unseen data streams whenever the accuracy drops below 90\%.  Consequently, our continual learning setup involved 13 generative models, as detailed in Table \ref{tab:CL_results}. Further details regarding the continual learning dataset are provided in Table \ref{tab:dataset-table}. Each subsequent task within this phase was trained for 5 epochs using the Adam optimizer with a learning rate of 1e-4. Our proposed data augmentation chain, consisting of RandomAffine (degrees=30, translate=0.1), RandomFlip (horizontal and vertical), and RandomAugment (N=3, M=9), was applied during continual training. The weight $\lambda$ and $\gamma$ were empirically set to 0.2 and 0.5, respectively. We also used a fixed replay set size of 500 samples.

\noindent
\textbf{Evaluation Metrics.} We assessed the generalization capability of AI-generated detectors using classification accuracy and average precision as metrics, following \cite{wang2020cnn, tan2023learning, ojha2023towards}. To evaluate the performance of our continual learning method, we utilized two primary metrics: average accuracy (AA) and average forgetting (AF). These metrics evaluate the model's ability to learn new tasks and its capacity to retain knowledge from previous tasks. We calculated these metrics using a test accuracy matrix $B \in \mathbb{R}^{t \times t}$, where each element $B_{i,j}$ represents the test accuracy on the $j$-th task after training on the $i$-th task. Here, $t$ denotes the total number of tasks involved. The Average Accuracy (AA) is defined as $AA = \frac{1}{t} \sum_{i=1}^t B_{t,i}$, reflecting the model’s overall performance at the end of the training sequence. Average Forgetting (AF) is calculated as $AF = \frac{1}{t-1} \sum_{j=1}^{t-1} (B_{t,j} - B_{j,j})$, measuring the decline in performance on earlier tasks as new ones are learned. 

\noindent
\textbf{Baselines.} We compared the cross-generator generalization performance of our method with that of several state-of-the-art baselines: (1) \textbf{CNNSpot}\cite{wang2020cnn}, which adopts a ResNet-50\cite{he2016deep} model (pre-trained on ImageNet\cite{deng2009imagenet}) and applies Gaussian blur and JPEG compression on ProGAN/LSUN images. (2) \textbf{FreDect}\cite{{frank2020leveraging}}, which leverages frequency abnormality of fake images. (3) \textbf{GramNet}\cite{liu2020global}, which incorporates global texture extraction into ResNet structure. (4) \textbf{LNP}\cite{liu2022detecting}, which identifies fake images from the frequency domain of the noise pattern. (5) \textbf{LGrad}\cite{tan2023learning}, which leverages gradient maps extracted from a well-trained image classifier, serving as the image's fingerprint. (6) \textbf{UnivFD}\cite{ojha2023towards}, which employs the pre-trained visual feature of CLIP\cite{radford2021learning} via nearest neighbor and linear probing. We employed its most effective variant for comparison, \textit{i.e.}  the linear probing classifier for CLIP ViT-L/14. (7) \textbf{NPR}\cite{tan2024rethinking}, which proposes an artifact representation focusing on local up-sampling artifacts in image pixels. (8) \textbf{FatFormer}\cite{liu2024forgery}, which incorporates a language-guided alignment strategy, leveraging contrasts between adapted image features and text prompt embeddings to further improve generalization. (9) \textbf{SAFE}\cite{li2024improving}, which employs three simple image transformations to alleviate training biases and enhances local awareness. For fair comparison, all these methods were uniformly trained on the ProGAN data from \cite{wang2020cnn}.

To evaluate the effectiveness of our continual learning method, we compared it against several established baselines, all utilizing the same initial offline model (ViT-B/32) proposed in our first stage: (1) \textbf{Sequential (Seq)} trains on a sequence of tasks; (2) \textbf{Joint} trains from scratch with all accumulated historical data; (3) \textbf{Experience Replay (ER)}\cite{chaudhry2019tiny} randomly selects a few old samples (10\% of the previous dataset in our experiments) into the reply set; (4) \textbf{Elastic Weight Consolidation (EWC)}\cite{kirkpatrick2017overcoming} constrains important parameters to stay close to their previous values, quantifying changes with the Fisher information matrix; (5) \textbf{Online Structured Laplace approximation (OSLA)}\cite{ritter2018online} approximates the diagonal blocks of the Hessian matrix with randomly selected data; (6) \textbf{Average Gradient Episode (A-GEM)}\cite{chaudhry2018efficient} maintains an average gradient matrix from historical data; (7) \textbf{Synaptic Intelligence (SI)}\cite{zenke2017continual} accumulates task-relevant information over time via intelligent synapses to determine the importance of parameters. (8) \textbf{Incremental Classifier and Representation Learning (iCaRL)}\cite{rebuffi2017icarl} uses distillation loss to ensure that the updated network's predictions remain similar to those of the network trained on the previous task.

\subsection{Generalization across Unseen Generators.}

We initially developed a foundational offline model designed to achieve high generalization capabilities. In this section, we evaluated its performance in cross-generator generalization by comparing it against several state-of-the-art methods. Table \ref{tab:generalization-gan} presents the classification accuracy and average precision results on various generative models, including GANs, deepfakes, and diffusion models. 

It can be observed that while previous methods demonstrate some cross-generator generalization, their performance often degrades significantly when faced with unknown generators exhibiting substantial architectural differences. For instance, CNNspot \cite{wang2020cnn} and LGrad \cite{tan2023learning} achieve impressive results on GANs (93.10\% mAP and 91.79\% mAP, respectively) but experience a noticeable drop in performance on diffusion models (61.42\% mAP and 60.18\% mAP, respectively). Recent work like UnivFD \cite{ojha2023towards} maintains consistently high performance across many tested scenarios, demonstrating the effectiveness of leveraging pre-trained vision-language models with unbiased features for AI-generated image detection. However, these studies typically rely on pre-trained models with frozen parameters, which limits their ability to learn forgery-specific features. In contrast, our method employ meticulous and efficient fine-tuning strategies, achieving substantial improvements over UnivFD with notable accuracy gains. 

\input{Table/table_4}
\begin{figure*}[h]
  \centering
  \includegraphics[width=\textwidth]{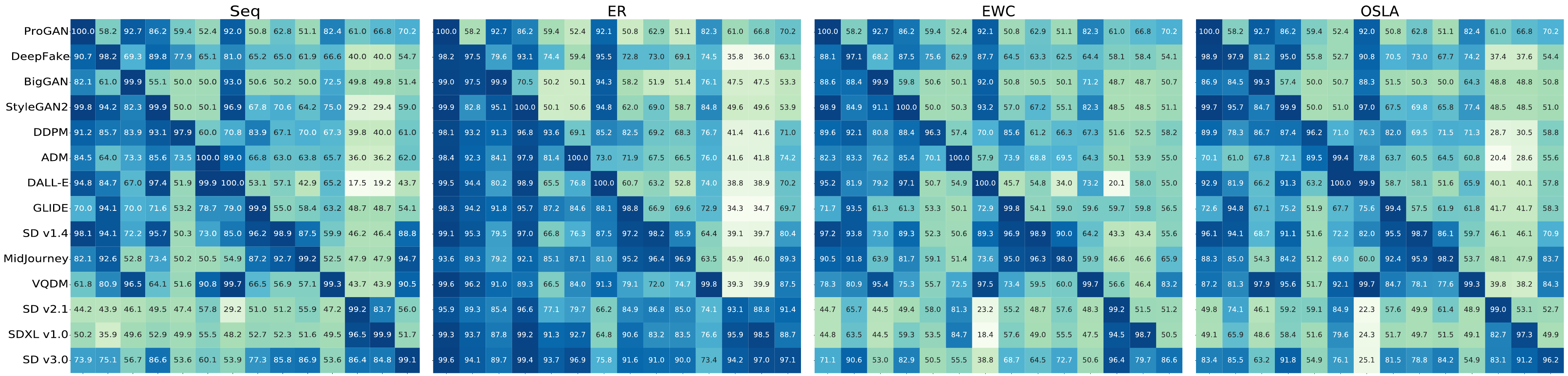}
  \includegraphics[width=\textwidth]{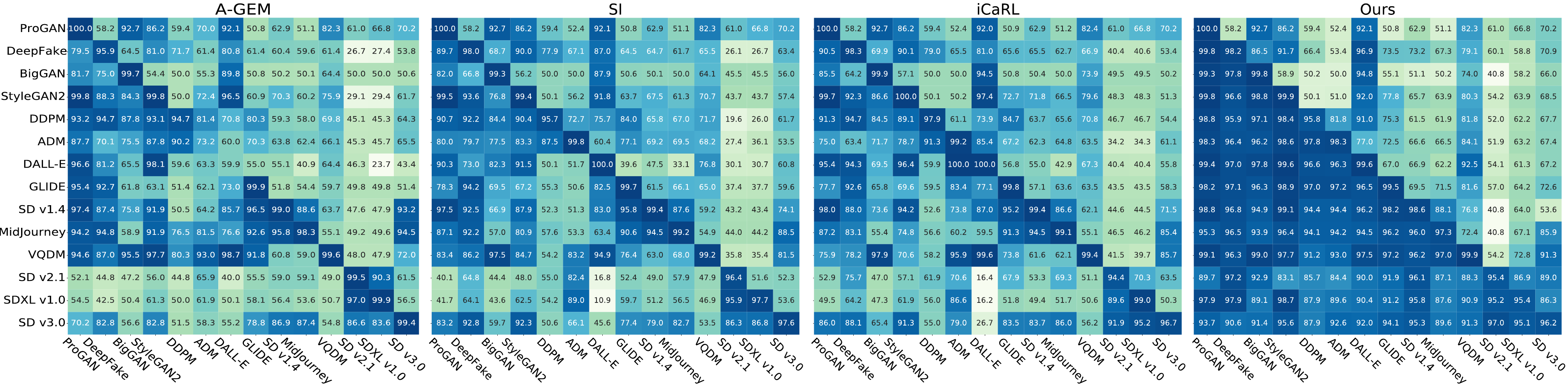}
  \caption{\textbf{Task-wise Continual Learning Performance.} The heatmap displays the test accuracy (ACC) for each task (x-axis) evaluated at the end of each sequential learning task (y-axis).}
  \label{fig:fig_3}
\end{figure*}

Notably, recent methods like NPR \cite{tan2024rethinking}, FatFormer \cite{liu2024forgery}, and SAFE \cite{li2024improving} show relatively robust detection on images from most generators. However, their performance significantly deteriorate on images generated by SD v2.1, SDXL v1.0, SD-Turbo, and SDXL-Turbo, achieving an average precision (AP) of approximately 50\%. This decline is primarily due to limited robustness to distribution shifts introduced by JPEG compression. Specifically, NPR \cite{tan2024rethinking} is designed to detect upsampling artifacts, SAFE \cite{li2024improving} enhances sensitivity to local textures through targeted data augmentations, and FatFormer \cite{liu2024forgery} incorporates both low-level texture cues and frequency-domain information via convolutional encoding and wavelet-based attention mechanisms. These methods share a common reliance on low-level artifacts, and consequently tend to exhibit reduced generalization and robustness when such features are suppressed or altered by post-processing techniques like JPEG compression. In contrast, models such as UnivFD \cite{ojha2023towards} and our method focus more on high-level semantic cues by leveraging vision-language models like CLIP to reason about semantic coherence. Although these approaches may exhibit slightly lower generalization performance compared to methods that rely on low-level artifacts, they tend to be more robust to compression-induced degradation and thereby achieve a more favorable trade-off between generalization and robustness. 

This vulnerability underscores the necessity of continual learning strategies that can equip detection models to keep pace with the rapidly evolving landscape of generative models. While our method may not achieve the highest performance on every individual generator, it consistently demonstrates strong average precision across all evaluated models. These results establish our method as a robust offline backbone with superior cross-generator generalization, laying a solid foundation for all subsequent continual learning experiments.

\subsection{Comparison with Existing Continual Learning Methods}
\begin{figure*}[t!]
  \centering
  \includegraphics[width=\textwidth]{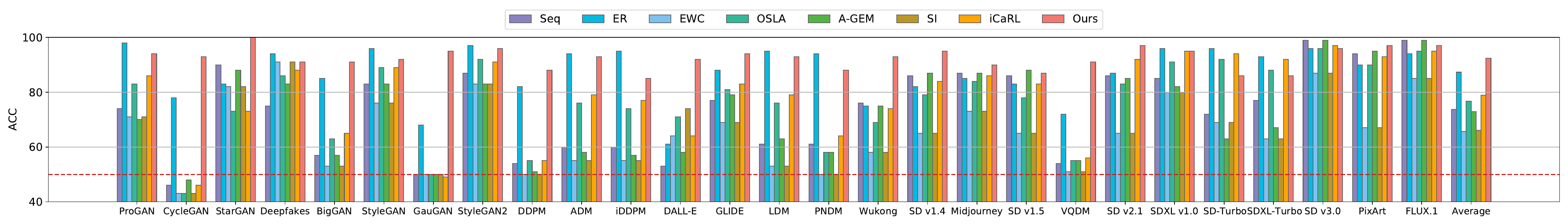}
  \includegraphics[width=\textwidth]{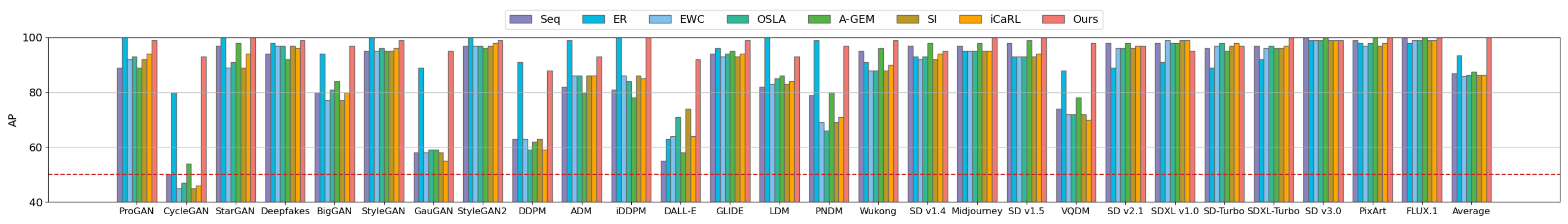}
  \caption{\textbf{Continual Learning Performance on Proposed Benchmark.} We evaluated the performance using ACC and AP metrics across all benchmark datasets after the final round of continual learning. The red dashed line represents the threshold of 50\% for both ACC and AP.}
  \label{fig:fig_5}
\end{figure*}

In the continual learning experiments, newly generated data was introduced as a sequence of tasks. We compared our proposed method with several established continual learning baselines, with detailed results presented in Table \ref{tab:CL_results}. Existing continual learning methods often face challenges in maintaining consistent performance on previous tasks, particularly when adapting to images generated by different generative families, such as transitioning from GANs to diffusion models. For example, the ER method experienced a significant decrease in AA, dropping from 96.96\% to 90.02\% (-6.94\% AA) when trained sequentially on StyleGAN2 and DDPM tasks. Similarly, the OSLA method saw a decrease from 94.98$\%$ AA to 87.69\% AA (-7.29\% AA). To provide a comprehensive understanding of how various methods adapt to learning new tasks while mitigating catastrophic forgetting, we conducted a task-wise performance analysis. This analysis is depicted through the diagonal of the heatmap in Figure \ref{fig:fig_3}, showing how performance evolves as the model sequentially learns new tasks. We observed that existing methods tend to overfit to new tasks due to the limited number of training samples available. All these methods, with the exception of A-GEM, achieved nearly 100\% detection performance on new tasks. However, their ability to preserve previously learned knowledge was compromised, as indicated by their forgetting metrics (AF) approaching 20\%. Additionally, we illustrated the results after the final round of continual learning across all benchmark datasets, as shown in Figure \ref{fig:fig_5}. All these comparison results demonstrate that our method exhibits superior plasticity and stability, achieving the best performance after training on all tasks. Specifically, our method employs an augmentation chain that progressively increases in complexity to preventing overfitting and learning new knowledge. Additionally, we utilized the K-FAC\cite{martens2015optimizing} method to approximate the Hessian matrix during updates, which significantly enhances the model's ability to preserve knowledge from previous tasks without compromising the learning of new information. Importantly, to retain the crucial commonalities needed for effective generalization across various generative models, as well as to learn the unique forgery traces of new generative models, we model the relationship between the old and new models based on linear mode connectivity. This approach allows us to achieve a trade-off between plasticity and stability, facilitating effective continual learning across varied tasks.

\begin{figure}[t]
  \centering
  \includegraphics[width=0.9\linewidth]{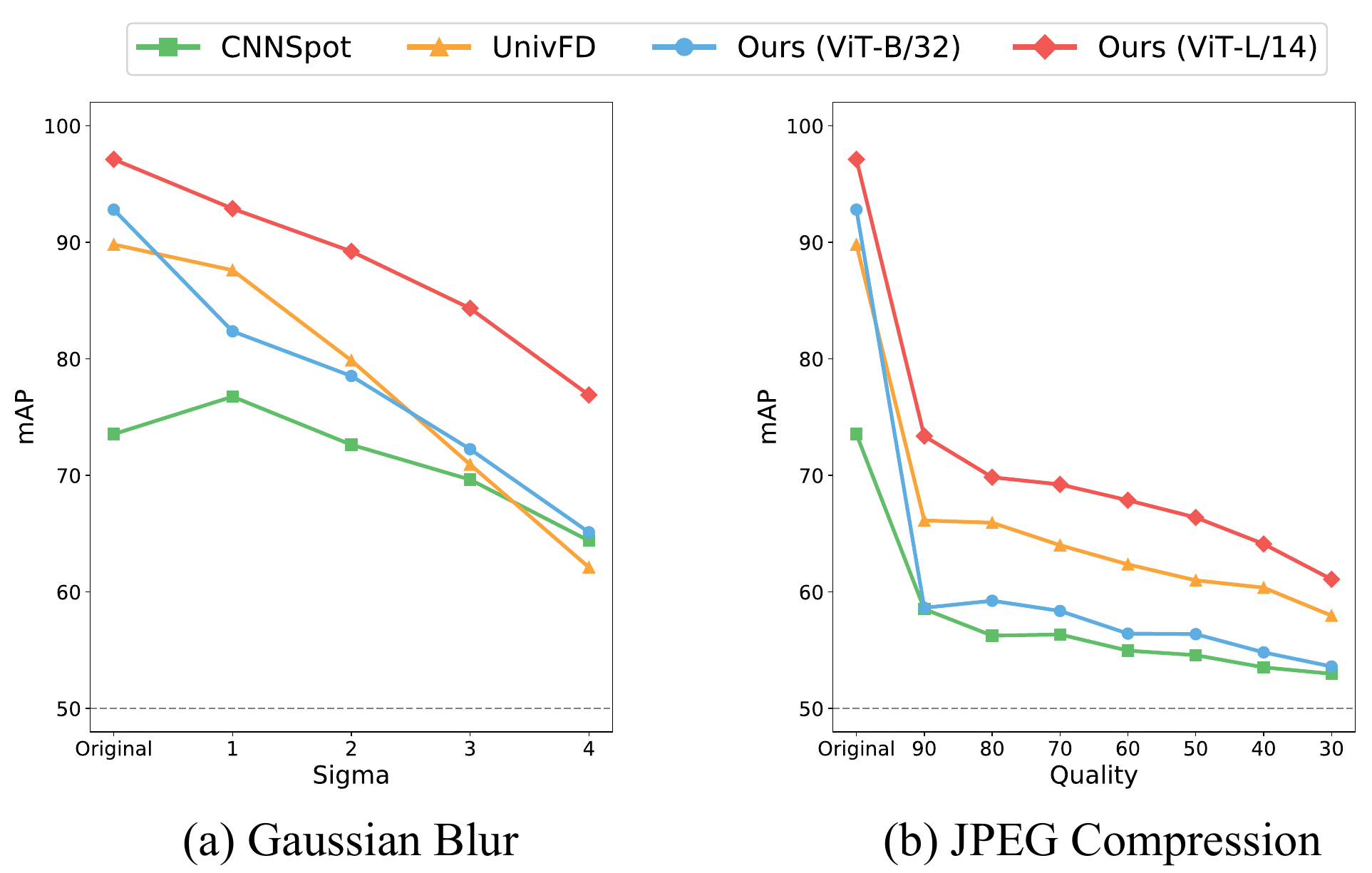}
  \caption{\textbf{Robustness to Post-processing Operations.} We compared the robustness of our method with CNNSpot\cite{wang2020cnn} and UnivFD\cite{ojha2023towards} against two test-time perturbations: (a) Gaussian Blur and (b) JPEG compression. `Original' denotes the results obtained from the original images without any post-processing operations.}
  \label{fig:fig_7}
\end{figure}

\subsection{Robustness to Post-processing Operations}
In real-world scenarios, images spread on public platforms often undergo various unknown degradations. Therefore, the robustness of detectors against common post-processing operations is crucial for practical applications. To assess this, we evaluated our method on two common post-processing operations: Gaussian blur and JPEG compression. Specifically, Gaussian blur was applied at four levels ($\sigma$ = 1, 2, 3, 4), and JPEG compression is applied at seven quality factors (QF = 90, 80, 70, 60, 50, 40, 30). We compared the robustness of our method with CNNSpot \cite{wang2020cnn} and UnivFD \cite{ojha2023towards}. Figure \ref{fig:fig_7} presents the mean Average Precision (mAP) results averaged across the 20 generators detailed in Table \ref{tab:dataset-table}. Our results show that all methods experience decreased performance when faced with Gaussian blur and JPEG compression. This decline can be attributed to the elimination of low-level artifacts that are crucial for distinguishing fake images from real ones. In contrast, our approach demonstrates better overall robustness compared to existing baselines, maintaining higher mAP scores under these perturbations. 

\subsection{Ablation Study}

\noindent
\textbf{Effect of Augmentation Chain (AC).} A series of studies was conducted to evaluate the impact of the proposed augmentation chain and linear interpolation on the continual learning process, as detailed in Table \ref{tab:ablation-study}. The results demonstrate that explicitly increasing the diversity and complexity of the training data significantly improves continual learning performance. Notably, the augmentation chain proves beneficial even with a limited dataset, such as 600 newly generated samples from GLIDE, leading to enhanced average accuracy. Additionally, following the implementation of continual learning using SD v2.1, the integration of the augmentation chain not only sustained higher average accuracy but also maintained consistent forgetting rates. These results suggest that progressively increasing complexity effectively prevents overfitting to a small set of new data, thereby ensuring stable knowledge retention across previously learned tasks. 

\input{Table/table_5}

\begin{figure*}[t!]
  \centering
  \includegraphics[width=\textwidth]{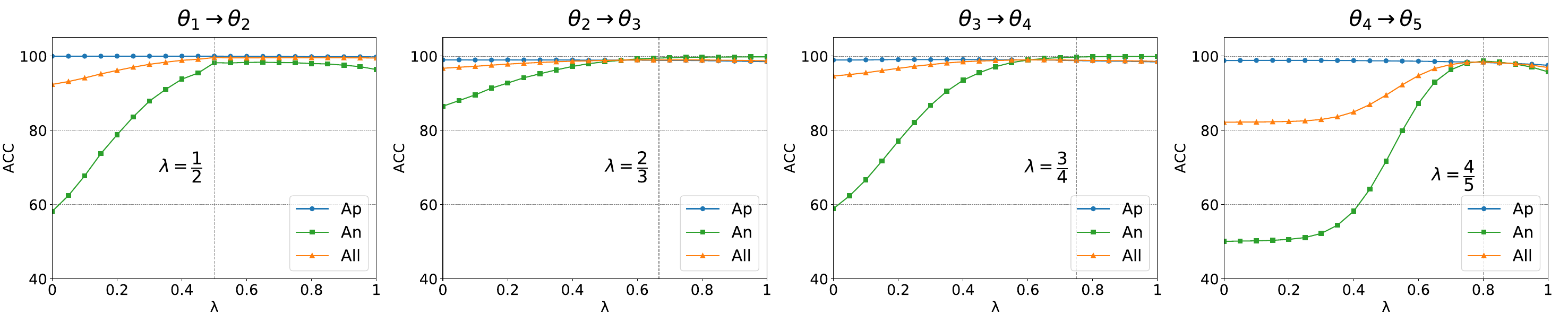}
  \caption{\textbf{Effect of Linear Interpolation.} Testing accuracy curves demonstrate the effect of linear interpolation between two adjacent minima on the accuracy of historical tasks (Ap), current tasks (An), and a combined set of all tasks (All). $\lambda$ represents the interpolation factor.}
  \label{fig:fig_8}
\end{figure*}

\noindent
\textbf{Effect of Linear Interpolation (Linear).} As demonstrated in Table \ref{tab:ablation-study}, the incorporation of the linear interpolation significantly mitigates the forgetting of previously learned knowledge enhances memorization capabilities. This finding supports the rationale and design of our methodology, where involves interpolating between adjacent minima and traversing along a linear path characterized by low errors. This approach effectively balances the trade-off between plasticity and stability, ensuring that the model not only retains prior knowledge but also integrates new information efficiently. Further exploration of the impact of $\lambda$, which adjusts the balance between networks from previously learned and current tasks, is illustrated in Figure \ref{fig:fig_8}.  It can be observed that as $\lambda$ increases from 0 to 1, there is a gradual decline in the ability of Ap to preserve previous knowledge, while An shows improved performance on the current tasks. Moreover, the model achieves a promising balance at $\lambda = \frac{T-1}{T}$ across all tasks, confirming our analysis. This suggests that the linear interpolation strategy is effective for continual learning in AI-generated image detection scenarios, balancing both knowledge retention and new information acquisition.

\noindent
\textbf{Effect of Replay.} As shown in Table \ref{tab:ablation-study}, we can observe that our proposed augmentation chain and linear interpolation significantly boost the performance of continual learning with previously learned knowledge preserved and new knowledge acquired, surpassing existing baseline methods. Furthermore, we found that introducing a replay set with a total size of only 500 can further enhance the performance of continual learning. Notably, the replay strategy does not require any additional design, simply selecting samples randomly is sufficient to achieve these improvements.

%% file: Table/table_1.tex
\begin{table}
\caption{\textbf{Continual learning dataset.} CL indicates the number of training samples used for each continual learning task.}

\label{tab:dataset-table}
\centering
\resizebox{\linewidth}{!}{  
\begin{tabular}{llcccccccccc}
\toprule  
\multirow{2}{*}{\shortstack[l]{Family}} & \multirow{2}{*}{Method} & \multirow{2}{*}{\shortstack[c]{Src}} & \multirow{2}{*}{\shortstack[c]{Format}} & \multicolumn{2}{c}{Ordering} & \multicolumn{5}{c}{Dataset size} \\ 

\cmidrule(lr){5-6} \cmidrule(lr){7-11} 

& &  & & $\#$ & Date & ToTal & Train & Val & CL & Test \\

\midrule

\multirow{7}{*}{\shortstack[l]{GAN}} & ProGAN~\cite{karras2018progressive} & ~\cite{wang2020cnn}  & PNG & 1 & Oct 17 & 368,060 & 360,060 & 4,000 & - & 4,000 \\ 

& CycleGAN~\cite{zhu2017unpaired} & ~\cite{wang2020cnn}  & PNG & 2 & Nov 17 & 1,321 & - & - & 132 & 1,189\\

& StarGAN~\cite{choi2018stargan} & ~\cite{wang2020cnn}  & PNG & 3 & Nov 17 & 1,999 & - & - & 200 & 1,799\\

& BigGAN~\cite{brock2019large} & ~\cite{wang2020cnn}  & PNG & 5 & Sep 18 & 2,000  & - & - & 200 & 1,800\\

& StyleGAN~\cite{karras2019style} & ~\cite{wang2020cnn}  & PNG & 6 & Dec 18 & 5,991 & - & - & 599 & 5,392 \\

& GauGAN~\cite{park2019semantic} & ~\cite{wang2020cnn}  & PNG & 7 & Mar 19 & 5,000 & - & - & 500 &  4,500\\

& StyleGAN2~\cite{karras2020analyzing} & ~\cite{wang2020cnn}  & PNG & 8 & Dec 19 & 7,988 & - & - & 799 & 7,189\\

\cdashline{1-11}

DeepFake & Deepfake & ~\cite{wang2020cnn}  & PNG & 4 & Nov 17 &  2,698 & - & - & 270 & 2,428 \\ 

\cdashline{1-11}

\multirow{19}{*}{\shortstack[l]{Diffusion}} & DDPM~\cite{ho2020denoising} & ~\cite{ricker2022towards}  & PNG & 9 & Jun 20 & 10,000 & - & - & 1,000 & 9,000 \\ 

& ADM~\cite{dhariwal2021diffusion} & ~\cite{ricker2022towards} & PNG & 10 & May 21 & 10,000 & - & - & 1,000 & 9,000 \\

& iDDPM~\cite{nichol2021improved} & ~\cite{ricker2022towards} & PNG & 11 & Dec 21 & 10,000 & - & - & 1,000 & 9,000\\ 

& DALL$\cdot$E ~\cite{Murahari2021DALLECI} & ~\cite{ojha2023towards}  & PNG & 12 & Dec 21 & 1,000 & - & - & 100 & 900 \\

& GLIDE~\cite{nichol2021glide} & ~\cite{zhu2024genimage} & PNG & 13 & Dec 21 & 168,000 & 162,000 & - & 600 & 6,000 \\

& LDM~\cite{rombach2022high}  & ~\cite{ricker2022towards}  & PNG & 14 & Dec 21 & 10,000 & - & - & 1,000 & 9,000\\

& PNDM~\cite{liu2022pseudo} & ~\cite{ricker2022towards} & PNG & 15 & Feb 22 & 10,000 & - & - & 1,000 & 9,000 \\

& Wukong~\cite{gu2022wukong} & ~\cite{zhu2024genimage} & PNG & 16 & Feb 22 & 168,000 & 162,000 & - & 600 & 6,000 \\

& SD v1.4~\cite{rombach2022high} & ~\cite{zhu2024genimage}  & PNG & 17 & Apr 22 & 168,000 & 162,000 & - & 600 & 6,000 \\

& Midjourney~\cite{midjourney} & ~\cite{zhu2024genimage} & PNG & 18 & Jul 22 & 168,000 & 162,000 & - & 600 & 6,000 \\

& SD v1.5~\cite{rombach2022high} & ~\cite{zhu2024genimage}  & PNG & 19 & Aug 22 & 174,000 & 166,000 & - & 800 & 8,000 \\

& VQDM~\cite{gu2022vector} & ~\cite{zhu2024genimage} & PNG & 20 & Aug 22 & 168,000 & 162,000 & - & 600 & 6,000 \\

& SD v2.1~\cite{rombach2022high} & ~\cite{chen2024drct} & JPEG & 21 & Nov 22 & 123,287 & 118,287 & - & 500 & 5,000 \\

& SDXL v1.0~\cite{podell2024sdxl} & ~\cite{chen2024drct} & JPEG & 22 & Jul 23 & 123,287 & 118,287 & - & 500 & 5,000 \\

& SD-Turbo~\cite{sauer2024fast} & ~\cite{chen2024drct} & JPEG & 23 & Nov 23 & 123,287 & 118,287 & - & 500 & 5,000 \\

& SDXL-Turbo~\cite{sauer2025adversarial} & ~\cite{chen2024drct} & JPEG & 24 & Nov 23 & 123,287 & 118,287 & - & 500 & 5,000 \\

& SD v3.0~\cite{esser2024scaling} & ~\cite{Li2024ImprovingSI} & PNG & 25 & Feb 24 & 5,000 & - & - & 500 & 4,500 \\

& PixArt-$\Sigma$~\cite{chen2024pixart} & ~\cite{Li2024ImprovingSI} & PNG & 26 & Mar 24 & 5,000 & - & - & 500 & 4,500 \\

& FLUX.1~\cite{FLUX} & ~\cite{Li2024ImprovingSI} & PNG & 27 & Aug 24 & 5,000 & - & - & 500 & 4,500 \\

\bottomrule
\end{tabular}
}
\end{table}

%% file: Table/table_2.tex
\begin{table*}[htbp]
\caption{\textbf{Generalization Comparisons on Benchmark Datasets.} All these methods are trained on ProGAN\cite{wang2020cnn} and evaluated using the Accuracy and average precision (ACC/AP) metrics. The best results are highlighted in bold, and the second-best is underlined.}

\label{tab:generalization-gan}
\centering
\resizebox{\linewidth}{!}{
\begin{tabular}{cccccccccccc}
\toprule
   Generator 
   & CNNSpot\cite{wang2020cnn}
   & FreDect\cite{frank2020leveraging} & GramNet\cite{liu2020global}
   & LGrad\cite{tan2023learning}
   & LNP\cite{liu2022detecting}
   & UnivFD\cite{ojha2023towards}
   & NPR\cite{tan2024rethinking}
   & FatFormer\cite{liu2024forgery}
   & SAFE\cite{li2024improving} 
   & Ours (ViT-B/32) 
   & Ours (ViT-L/14) \\

    \midrule
    ProGAN & 99.74 / \textbf{100.0} & 99.36 / \underline{99.99} & \underline{99.99} / \textbf{100.0} & 99.78 / \textbf{100.0} & 99.71 / \textbf{100.0} & 99.81 / \textbf{100.0} & 99.96 / \textbf{100.0} & \textbf{100.0} / \textbf{100.0} & 99.98 / \textbf{100.0} & \underline{99.99} / \textbf{100.0} & \textbf{100.0} / \textbf{100.0} \\
    
    \midrule
    StyleGAN & 84.64 / 99.36 & 78.02 / 88.98 & 83.59 / 94.49 & 89.63 / 98.03 & 91.20 / 97.37 & 80.40 / 97.48 & \underline{99.25} / \textbf{100.0} & \textbf{99.42} / \underline{99.99} & 93.24 / 99.96 & 91.66 / 98.56 & 94.59 / 99.97 \\
    
    \midrule
    BigGAN & 69.30 / 81.89 & 81.97 / 93.62 & 67.55 / 62.34 & 81.73 / 89.08 & 81.80 / 90.97 & 94.50 / 99.27 & 76.85 / 83.10 & \underline{96.83} / \underline{99.82} & 90.25 / 95.71 & 92.68 / 98.94 & \textbf{99.68} / \textbf{99.99} \\
    
    \midrule
    CycleGAN & 82.85 / 90.07 & 78.77 / 84.78 & 73.73 / 74.82 & 85.96 / 93.84 & 83.54 / 94.06 & \underline{98.33} / 99.46 & 90.27 / 97.63 & 95.87 / \underline{99.85} & 98.22 / 99.71 & 97.05 / 99.61 & \textbf{99.62} / \textbf{99.99} \\
    
    \midrule
    StarGAN & 90.72 / 98.09 & 94.62 / 99.49 & 95.05 / \textbf{100.0} & 98.10 / 99.98 & \textbf{99.90} / \textbf{100.0} & 95.75 / 99.37 & 97.50 / 99.51 & 98.12 / \textbf{100.0} & \underline{99.92} / \textbf{100.0} & 99.37 / \textbf{100.0} & 99.85 / \textbf{100.0} \\
    
    \midrule
    GauGAN & 77.68 / 87.64 & 80.57 / 82.86 & 57.78 / 55.28 & 80.08 / 91.32 & 72.48 / 79.07 & \underline{99.47} / \underline{99.98} & 78.70 / 79.29 & 94.36 / 99.84 & 86.09 / 94.38 & 87.58 / 99.57 & \textbf{99.95} / \textbf{100.0} \\
    
    \midrule
    StyleGAN2 & 80.88 / 98.75 & 66.17 / 82.53 & 85.86 / 99.37 & 86.11 / 98.26 & 93.09 / 99.20 & 70.76 / 97.71 & \underline{98.41} / \textbf{99.97} & \textbf{99.22} / 99.89 & 98.92 / \underline{99.94} & 86.25 / 99.06 & 93.60 / 99.72 \\
    
    \midrule
    Deepfake & 53.47 / 89.02 & 63.29 / 70.77 & 63.13 / 95.30 & 51.88 / 63.83 & 55.30 / 76.04 & 68.57 / 81.76 & \textbf{89.71} / 95.24 & 84.10 / \underline{97.59} & \underline{89.14} / 96.98 & 72.19 / 91.34 & 77.69 / \textbf{98.06} \\
    
    \midrule
    Average & 79.91 / 93.10 & 80.35 / 87.88 & 78.95 / 85.20 & 84.16 / 91.79 & 84.63 / 92.09 & 88.45 / 96.88 & 91.33 / 94.34 & \underline{95.39} / \underline{99.62} & 95.22 / 98.34 & 90.85 / 98.34 & \textbf{95.63} / \textbf{99.72} \\

\bottomrule
\end{tabular}}

\end{table*}

%% file: Table/table_3.tex
\begin{table*}[htbp]
\centering
\resizebox{\linewidth}{!}{
\begin{tabular}{cccccccccccc}
\toprule
    Generator 
    & CNNSpot\cite{wang2020cnn} 
    & FreDect\cite{frank2020leveraging}
    & GramNet\cite{liu2020global} 
    & LGrad\cite{tan2023learning} 
    & LNP\cite{liu2022detecting} 
    & UnivFD\cite{ojha2023towards} 
    & NPR \cite{tan2024rethinking}
    & FatFormer\cite{liu2024forgery}
    & SAFE\cite{li2024improving}
    & Ours (ViT-B/32) 
    & Ours (ViT-L/14) \\
    \midrule
    DDPM & 50.69 / 74.63 & 62.51 / 64.74 & 50.04 / 67.87 & 55.17 / 62.56 & 51.85 / 71.04 & 67.32 / 95.79 & \textbf{99.29} / \textbf{100.0} & 55.58 / 63.10 & \underline{89.19} / 99.09 & 69.41 / 94.74 & 78.31 / \underline{99.13} \\
   
    \midrule
    ADM & 50.20 / 61.66 & 69.80 / 74.97 & 50.02 / 56.65 & 47.65 / 34.95 & 49.65 / 39.02 & 53.48 / 81.15 & \textbf{84.39} / \textbf{99.41} & 55.95 / 82.54 & \underline{79.41} / \underline{96.75} & 52.38 / 89.09 & 54.11 / 90.69 \\
    
    \midrule
    iDDPM & 50.48 / 72.60 & 74.15 / 78.38 & 50.02 / 64.08 & 49.41 / 45.24 & 49.58 / 40.37 & 62.11 / 92.19 & \textbf{95.95} / \textbf{99.94} & 64.69 / 87.26 & \underline{89.40} / 98.25 & 63.68 / 89.72 & 69.21 / \underline{98.42} \\
    
    \midrule
    DALL$\cdot$E & 57.10 / 83.17 & 81.55 / 94.98 & 87.80 / 98.81 & 85.75 / 95.60 & 83.60 / 94.90 & 87.30 / 94.47 & 86.05 / 98.54 & \underline{95.00} / 97.85 & \textbf{98.50} / \underline{99.59} & 92.10 / 99.11 & \underline{95.95} / \textbf{99.91} \\
    
    \midrule
    GLIDE & 51.15 / 59.88 & 52.80 / 51.77 & 55.15 / 53.46 & 67.00 / 75.95 & 64.40 / 70.02 & 57.85 / 83.78 & 81.70 / 91.46 & \textbf{92.10} / \textbf{97.88} & \underline{88.50} / \underline{95.10} & 58.45 / 74.76 & 70.95 / 85.13 \\
    
    \midrule
    LDM & 50.52 / 70.80 & 74.20 / 86.91 & 54.16 / 98.72 & 59.37 / 71.50 & 54.90 / 79.24 & 66.11 / 94.62 & \textbf{99.95} / \textbf{100.0} & 86.02 / 98.42 & \underline{99.94} / \textbf{100.0} & 69.27 / 92.68 & 76.37 / \underline{99.04} \\
    
    \midrule
    PNDM & 51.31 / 74.61 & 58.17 / 56.80 & 50.13 / 91.14 & 55.01 / 60.74 & \underline{85.17} / 98.37 & 73.95 / 97.31 & 50.94 / 99.33 & \textbf{99.46} / \underline{99.09} & 75.84 / 97.53 & 70.25 / 91.04 & 84.28 / \textbf{99.31} \\
    
    \midrule
    Wukong & 50.05 / 55.02 & 41.70 / 42.04 & 62.65 / 59.61 & 60.55 / 62.31 & 58.30 / 64.90 & 54.35 / 80.36 & 78.90 / 85.50 & \textbf{96.45} / \underline{96.89} & \underline{91.95} / 96.65 & 84.35 / 95.49 & 86.56 / \textbf{97.84} \\
    
    \midrule
    SD v1.4 & 52.20 / 56.58 & 38.75 / 38.34 & 65.65 / 62.74 & 64.40 / 63.12 & 61.65 / 68.59 & 51.85 / 68.11 & 83.65 / 90.23 & \textbf{95.40} / \textbf{99.20} & \underline{93.40} / \underline{98.78} & 84.45 / 94.66 & 86.90 / 97.10 \\
    
    \midrule
    Midjourney & 50.40 / 53.40 & 43.55 / 42.94 & 60.80 / 58.65 & 68.65 / 73.60 & 49.95 / 49.88 & 50.45 / 56.75 & \underline{82.40} / \underline{91.75} & 70.70 / 82.47 & \textbf{91.35} / \textbf{97.32} & 59.35 / 74.71 & 77.45 / 82.67 \\
    
    \midrule
    SD v1.5 & 52.20 / 58.06 & 39.10 / 38.37 & 65.15 / 62.86 & 64.20 / 64.12 & 61.10 / 68.23 & 51.95 / 65.82 & 82.75 / 90.38 & 83.90 / 96.24 & \textbf{92.30} / \textbf{98.83} & 82.75 / 93.87 & \underline{87.21} / \underline{97.05} \\
    
    \midrule
    VQDM & 53.05 / 55.02 & 78.40 / 85.77 & 59.15 / 55.18 & 69.85 / 73.65 & 64.50 / 70.15 & 85.25 / 97.34 & 72.05 / 78.38 & \textbf{94.00} / \underline{98.99} & 90.55 / 98.77 & 88.70 / 97.59 & \underline{91.95} / \textbf{99.18} \\

    \midrule
    SD v2.1 & 48.57 / 55.76 & 31.64 / 34.77 & 32.73 / 32.99 & 31.45 / 33.17 & 42.29 / 36.66 & \underline{57.26} / \underline{85.36} & 49.45 / 42.20 & 51.80 / 51.34 & 49.94 / 51.22 & 60.14 / 82.58 & \textbf{60.98} / \textbf{86.36} \\

    \midrule
    SDXL v1.0 & 49.55 / 64.64 & 59.05 / 59.02 & 32.72 / 33.14 & 31.52 / 32.23 & 40.30 / 34.76 & \underline{65.07} / \textbf{90.80} & 49.44 / 36.77 & 56.65 / 58.42 & 49.96 / 54.94 & 53.04 / 83.13 & \textbf{66.79} / \underline{84.97} \\

    \midrule
    SD-Turbo & 48.57 / 52.64 & \textbf{73.24} / \textbf{82.09} & 32.72 / 35.88 & 30.98 / 31.72 & 43.29 / 36.25 & \underline{56.64} / 77.56 & 49.46 / 39.08 & 54.30 / 58.63 & 49.92 / 50.94 & 59.42 / 73.36 & 51.29 / \underline{78.48} \\

    \midrule
    SDXL-Turbo & 49.75 / 64.03 & \textbf{74.44} / \textbf{81.73} & 32.32 / 33.18 & 30.96 / 31.36 & 40.60 / 31.86 & \underline{53.59} / 76.42  & 49.50 / 36.82 & 51.69 / 50.63 & 49.90 / 51.07 & 53.41 / 75.13 & 50.64 / \underline{76.67} \\

    \midrule
    SD v3.0 & 49.91 / 50.94 & 40.98 / 42.15 & 59.15 / 58.15 & 70.13 / 77.89 & 57.14 / 62.15 & 50.12 / 59.05 & \underline{82.95} / \underline{94.02} & 74.49 / 89.52 & \textbf{83.58} / \textbf{94.08} & 68.60 / 76.46 & 70.24 / 81.63 \\

    \midrule
    PixArt-Sigma & 48.52 / 50.78 & 39.45 / 40.84 & 59.56 / 58.55 & 67.40 / 68.17 & 45.05 / 41.28 & 50.78 / 65.90 & \underline{83.59} / \underline{93.77} & 82.98 / 92.80 & \textbf{84.76} / \textbf{94.31} & 75.24 / 77.89 & 71.08 / 82.03 \\

    \midrule
    FLUX.1 & 49.06 / 47.32 & 40.81 / 41.70 & 58.29 / 57.42 & 71.69 / 81.11 & 56.53 / 60.49 & 50.04 / 55.09 & \underline{82.92} / \underline{94.35} & 75.64 / 89.58 & \textbf{83.57} / \textbf{94.95} & 74.16 / 79.79 & 73.02 / 84.38 \\

    \midrule
    Avgerage & 50.96 / 61.42 & 54.34 / 54.38 & 53.29 / 60.36 & 57.88 / 60.18 & 56.66 / 60.36 & 60.84 / 79.17 & \underline{75.83} / 84.15 & 75.47 / 83.18 & \textbf{80.63} / \underline{87.79} & 69.39 / 84.07 & 74.55 / \textbf{89.66} \\
    
\bottomrule
\end{tabular}}
\label{tab:generalization-diffusion-transposed}
\end{table*}

%% file: Table/table_4.tex
\begin{table*}
\caption{\textbf{Comparative Continual Learning Performance.} All methods employed the same initial offline model, i.e., the CLIP ViT-B/32 version. Performance is evaluated based on average accuracy (AA) and average forgetting (AF).}
\label{tab:CL_results}
\centering
\resizebox{1\linewidth}{!}{
\begin{tabular}{c|c|cc|cc|cc|cc|cc|cc|cc|cc|cc|cc|cc|cc|cc} 
\hline
\multirow{2}{*}{Method} & ProGAN & \multicolumn{2}{c|}{Deepfake} & \multicolumn{2}{c|}{BigGAN} & \multicolumn{2}{c|}{StyleGAN2}  & \multicolumn{2}{c|}{DDPM} & \multicolumn{2}{c|}{ADM} & \multicolumn{2}{c|}{DALL$\cdot$E} & \multicolumn{2}{c|}{GLIDE} & \multicolumn{2}{c|}{SD v1.4} & \multicolumn{2}{c|}{Midjourney} & \multicolumn{2}{c}{VQDM} &
\multicolumn{2}{c|}{SD v2.1} &
\multicolumn{2}{c|}{SDXL v1.0} &
\multicolumn{2}{c}{SD v3.0} \\
\cline{2-28} & AA & AA & AF & AA & AF & AA & AF & AA & AF & AA & AF & AA & AF & AA & AF & AA & AF & AA & AF & AA & AF & AA & AF & AA & AF & AA & AF \\ 
\hline
Seq & 99.99 & 94.46 & -9.29 & 81.01 & -25.39 & 94.05 & -15.78 & 90.35 & -12.65 & 80.14 & -17.47 & 79.66 & -16.04 & 76.68 & -14.67 & 84.50 & -13.32 & 73.82 & -15.49 & 74.31 & -20.34 & 51.90 & -21.93 & 57.29 & -23.62 & 73.85 & -23.04 \\

\hline
Joint & 99.99 & 98.77 & -0.02 & 99.34 & -0.06 & 99.57 & -0.03 & 98.15 & -0.21 & 98.97 & -0.27 & 98.81 & -0.31 & 98.54 & -0.37 & 98.80 & -0.40 & 98.30 & -0.54 & 98.38 & -0.51 & 95.69 & -3.12 & 95.17 & -2.39 & 96.01 & -1.77 \\

\hline
ER & 99.99 & 98.26 & -1.15 & 96.77 & -1.00 & 96.96 & -0.71 & 90.92 & -0.54 & 85.97 & -0.16 & 86.18 & -1.11 & 84.65 & -1.85 & 84.11 & -1.68 & 83.94 & -1.19 & 83.73 & -1.52 & 84.51 & -12.09 & 86.02 & -12.15 & 90.68 & -12.10 \\

\hline
EWC & 99.99 & 92.57 & -11.92 & 91.63 & -11.17 & 93.73 & -9.43 & 89.44 & -10.86 & 82.87 & -15.46 & 79.85 & -21.60 & 70.47 & -24.89 & 82.35 & -21.05 & 80.11 & -19.52 & 77.10 & -23.16 & 67.97 & -35.08 & 69.39 & -26.50 & 78.11 & -21.52 \\

\hline
OSLA & 99.99 & 98.37 & -1.11 & 90.25 & -10.21 & 94.98 & -9.08 & 87.69 & -10.99 & 76.65 & -15.06 & 85.05 & -14.40 & 75.53 & -19.85 & 83.32 & -17.99 & 77.86 & -18.04 & 85.92 & -18.22 & 69.61 & -26.25 & 73.33 & -25.62 & 79.08 & -23.77 \\

\hline
A-GEM & 99.99 & 87.70 & -20.48 & 85.48 & -20.13 & 93.03 & -14.22 & 92.70 & -10.43 & 85.18 & -11.30 & 80.14 & -12.02 & 70.58 & -14.28 & 83.37 & -12.78 & 78.33 & -13.31 & 75.76 & -18.08 & 66.07 & -25.96 & 68.50 & -26.39 & 73.91 & -25.52 \\

\hline
SI & 99.99 & 93.82 & -10.32 & 82.67 & -22.70 & 92.32 & -16.64 & 90.68 & -12.76 & 84.65 & -13.40 & 76.98 & -22.22 & 74.66 & -23.23 & 80.73 & -20.94 & 77.57 & -20.46 & 80.98 & -22.40 & 64.58 & -25.77 & 69.54 & -27.71 & 75.26 & -17.40 \\ 

\hline
iCaRL & 99.99 & 94.38  & -9.51 & 83.19 & -23.06 & 94.63 & -13.82 & 91.52 & -12.00 & 79.87 & -14.57 & 87.93 & -13.70 & 78.18 & -17.95 & 84.64 & -16.50 & 76.18 & -17.72 & 79.38 & -21.15 & 64.26 & -25.79 & 65.29 & -27.34 & 70.51 & -24.72 \\

\hline
\textbf{Ours} & \textbf{99.99} & \textbf{99.06} & \textbf{-0.09} & \textbf{99.13} & \textbf{-0.21} & \textbf{99.07} & \textbf{-0.24} & \textbf{98.38} & \textbf{-0.50} & \textbf{97.30} & \textbf{-1.15} & \textbf{96.13} & \textbf{-1.43} & \textbf{95.95} & \textbf{-1.18} & \textbf{94.90} & \textbf{-1.44} & \textbf{93.53} & \textbf{-1.68} & \textbf{96.20} & \textbf{-1.56} & \textbf{87.41} & \textbf{-13.20} & \textbf{89.55} & \textbf{-11.39} & \textbf{92.32} & \textbf{-8.91} \\
\hline
\end{tabular}
}
\end{table*}

%% file: Table/table_5.tex
\begin{table*}
\caption{\textbf{Ablation study on the effect of our proposed modules.} Settings same as in Table \ref{tab:CL_results}.}
\label{tab:ablation-study}
\centering
\resizebox{1\linewidth}{!}{
\begin{tabular}{c|c|cc|cc|cc|cc|cc|cc|cc|cc|cc|cc|cc|cc|cc} 
\hline
\multirow{2}{*}{Method} & ProGAN & \multicolumn{2}{c|}{Deepfake} & \multicolumn{2}{c|}{BigGAN} & \multicolumn{2}{c|}{StyleGAN2}  & \multicolumn{2}{c|}{DDPM} & \multicolumn{2}{c|}{ADM} & \multicolumn{2}{c|}{DALL$\cdot$E} & \multicolumn{2}{c|}{GLIDE} & \multicolumn{2}{c|}{SD v1.4} & \multicolumn{2}{c|}{Midjourney} & \multicolumn{2}{c|}{VQDM} &
\multicolumn{2}{c|}{SD v2.1} &
\multicolumn{2}{c|}{SDXL v1.0} &
\multicolumn{2}{c}{SD v3.0} \\
\cline{2-28} & AA & AA & AF & AA & AF & AA & AF & AA & AF & AA & AF & AA & AF & AA & AF & AA & AF & AA & AF & AA & AF & AA & AF & AA & AF & AA & AF \\ 

\hline
K-FAC & 99.99 & 98.37 & -1.11 & 90.25 & -10.21 & 94.98 & -9.08 & 87.69 & -10.99 & 76.65 & -15.06 & 85.05 & -14.40 & 75.53 & -19.85 & 83.32 & -17.99 & 77.86 & -18.04 & 85.92 & -18.22 & 69.61 & -26.25 & 73.33 & -25.62 & 79.08 & -23.77 \\

\hline
K-FAC+AC & 99.99 & 98.49 & -1.15 & 97.43 & -2.08 & 93.12 & -6.78 & 88.17 & -9.73 & 81.63 & -14.68 & 86.47 & -14.41 & 77.41 & -16.77 & 85.21 & -17.29 & 79.78 & -17.90 & 87.72 & -17.70 & 73.27 & -16.82 & 75.63 & -15.24 & 81.41 & -16.39 \\

\hline
K-FAC+Linear & 99.99 & 98.62 & -0.20 & 98.62 & -0.14 & 97.88 & -0.47 & 93.40 & -3.22 & 89.32 & -6.42 & 90.01 & -6.04 & 90.08 & -5.67 & 90.27 & -4.98 & 90.25 & -4.38 & 90.51 & -4.00 & 77.49 & -14.33 & 82.10 & -15.41 & 85.96 & -12.96 \\

\hline
K-FAC+AC+Linear & 99.99 & 98.73 & -0.12 & 98.79 & -0.67 & 98.07 & -0.35 & 94.61 & -3.24 & 92.36 & -4.75 & 87.88 & -7.81 & 92.31 & -8.58 & 88.56 & -9.22 & 89.58 & -10.00 & 92.29 & -6.16 & 86.72 & -14.98 & 88.59 & -13.09 & 90.97 & -7.15 \\

\hline
\textbf{K-FAC+AC+Linear+Reply (Ours)} & \textbf{99.99} & \textbf{99.06} & \textbf{-0.09} & \textbf{99.13} & \textbf{-0.21} & \textbf{99.07} & \textbf{-0.24} & \textbf{98.38} & \textbf{-0.50} & \textbf{97.30} & \textbf{-1.15} & \textbf{96.13} & \textbf{-1.43} & \textbf{95.95} & \textbf{-1.18} & \textbf{94.90} & \textbf{-1.44} & \textbf{93.53} & \textbf{-1.68} & \textbf{96.20} & \textbf{-1.56} & \textbf{87.41} & \textbf{-13.20} & \textbf{89.55} & \textbf{-11.39} & \textbf{92.32} & \textbf{-8.91} \\
\hline
\end{tabular}
}
\end{table*}

%% file: Section/6_conclusion.tex
\section{Conclusion}
This paper aims to develop a practical method for AI-generated image detection. We propose a three-stage domain continual learning framework to continuously enhance generalization capabilities. The first stage employs a strategic parameter-efficient fine-tuning approach to develop a transferable offline detection model with high generalization capability. For new tasks, a carefully designed data augmentation chain with increasing complexity efficiently learns from a limited number of novel samples and prevents overfitting. Additionally, the Kronecker-Factored Approximate Curvature (K-FAC) method is used to mitigate catastrophic forgetting. Identifying and effectively leveraging shared characteristics across diverse feature distributions is crucial for achieving robust generalization across various generative models. The final stage employs a linear interpolation strategy with Linear Mode Connectivity to capture commonalities across diverse generative models, enhancing overall performance. Our extensive experiments across 27 generative models demonstrate superior generalization to unseen generators and robustness against common post-processing operations. Furthermore, the continual learning strategies effectively and continuously improve generalization, ensuring sustained effectiveness in evolving environments. 

%% file: Section/7_acknowledgement.tex
\section{ACKNOWLEDGEMENTS}
The work described in this paper was supported by the National Natural Science Foundation of China (Grant 62271307). We are grateful to the anonymous reviewers for their insightful comments and helpful suggestions, which greatly contributed to improving this work.

%% file: Section/8_authors.tex
\begin{IEEEbiography}[{\includegraphics[width=1in,height=1.25in,clip,keepaspectratio]{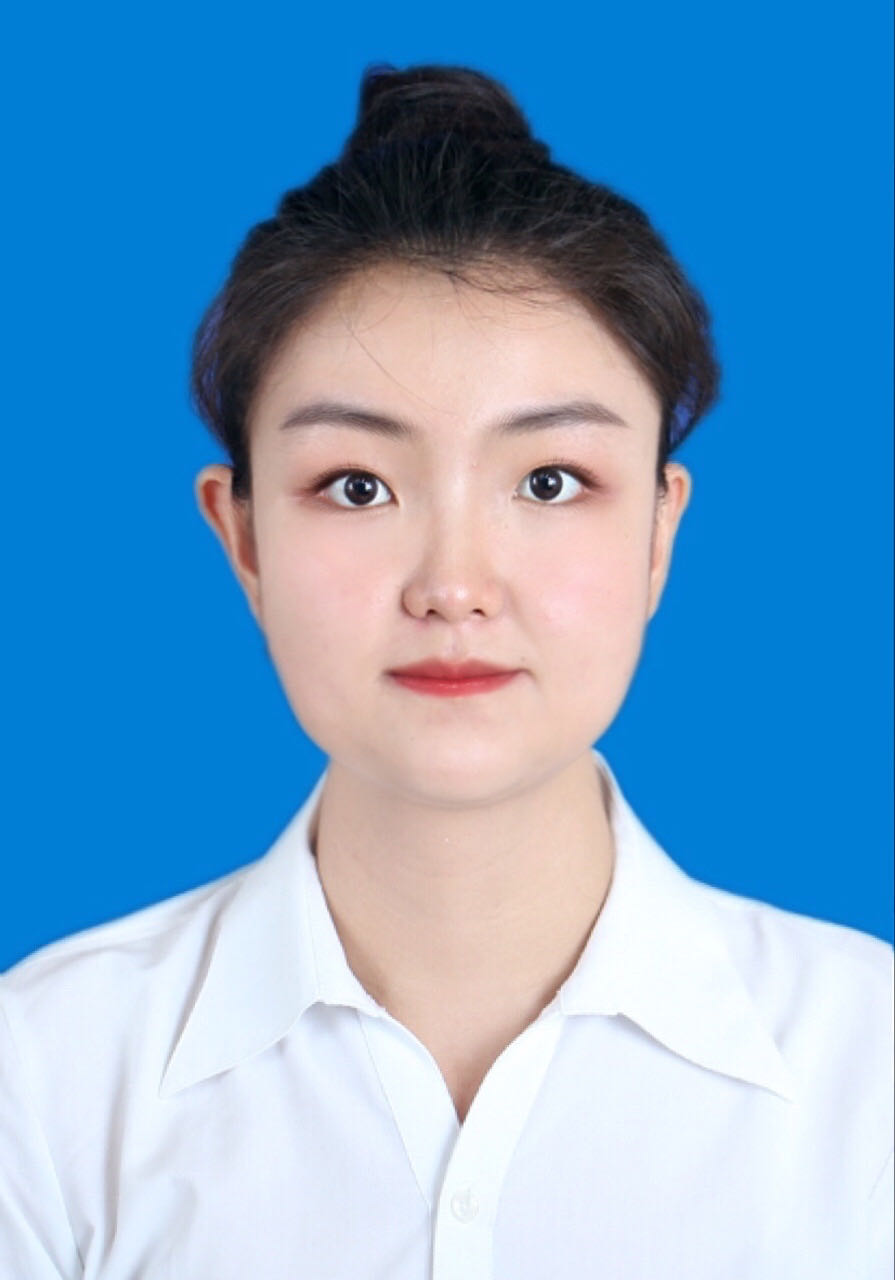}}]{Hanyi Wang} received the B.Eng. degree from Xidian University, China, in 2021. She is currently pursuing the Ph.D. degree with the School of Electric Information and Electronic Engineering, Shanghai Jiao Tong University, Shanghai, China. Her research interests include DeepFake detection and generative image forensics.\end{IEEEbiography}

\vspace{-1.5cm}

\begin{IEEEbiography}[{\includegraphics[width=1in,height=1.25in,clip,keepaspectratio]{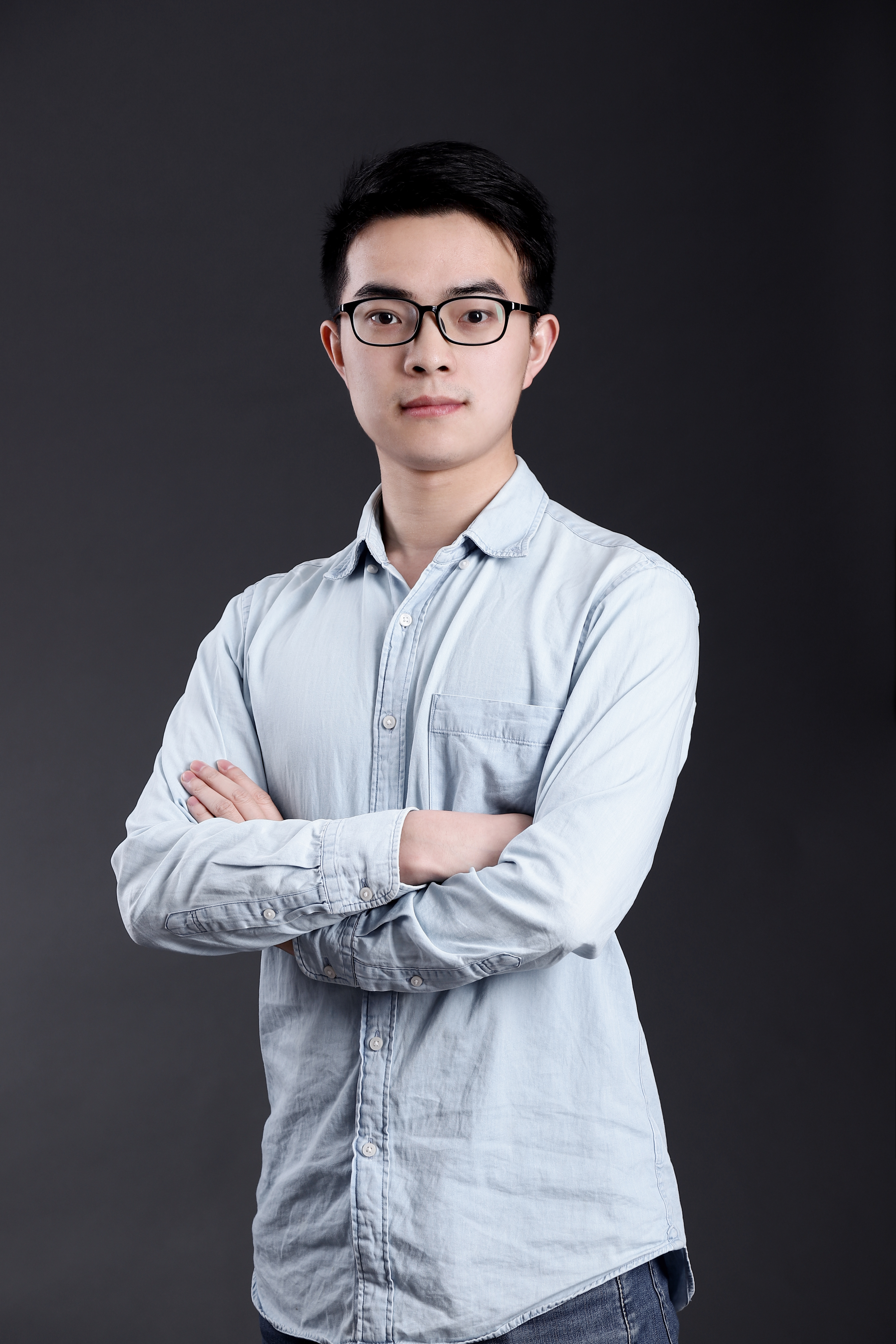}}]{Jun Lan} received the B.Eng. degree from Tianjin University, Tianjin, China, in 2014, and the M.Eng. degree from Shanghai Jiao Tong University, Shanghai, China, in 2017. He currently works as a Senior Algorithm Expert at Ant Group. His research interests include computer vision and pattern recognition.\end{IEEEbiography}

\begin{IEEEbiography}[{\includegraphics[width=1in,height=1.25in,clip,keepaspectratio]{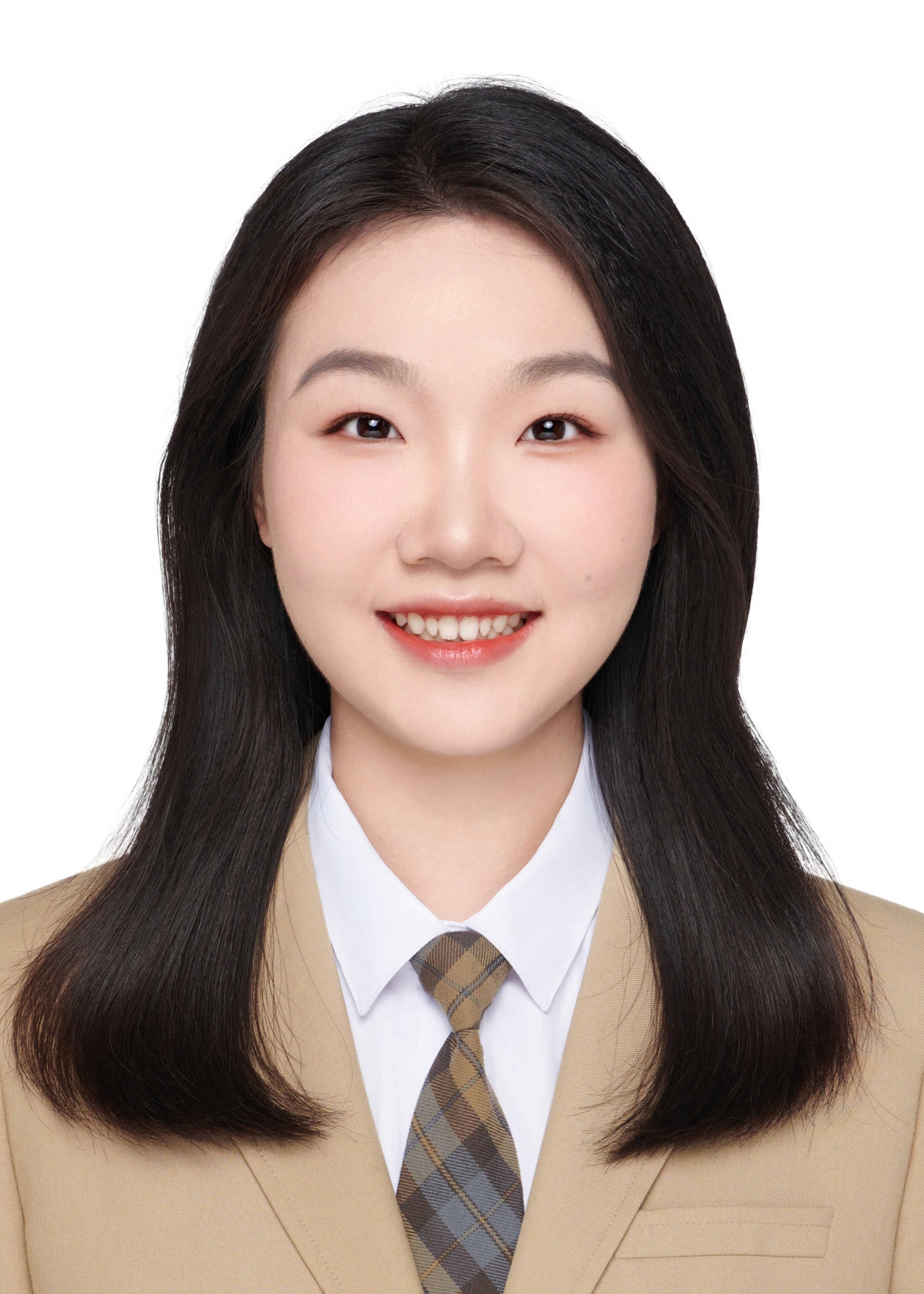}}]{Yaoyu Kang} received the B.Eng. degree from Shanghai Jiao Tong University, Shanghai, China, in 2025. She is going to pursue the M.Eng. degree with the School of Electric Information and Electronic Engineering, Shanghai Jiao Tong University, Shanghai, China. Her research interests include computer vision and pattern recognition.\end{IEEEbiography}

\vspace{-2.5cm}

\begin{IEEEbiography}[{\includegraphics[width=1in,height=1.25in,clip,keepaspectratio]{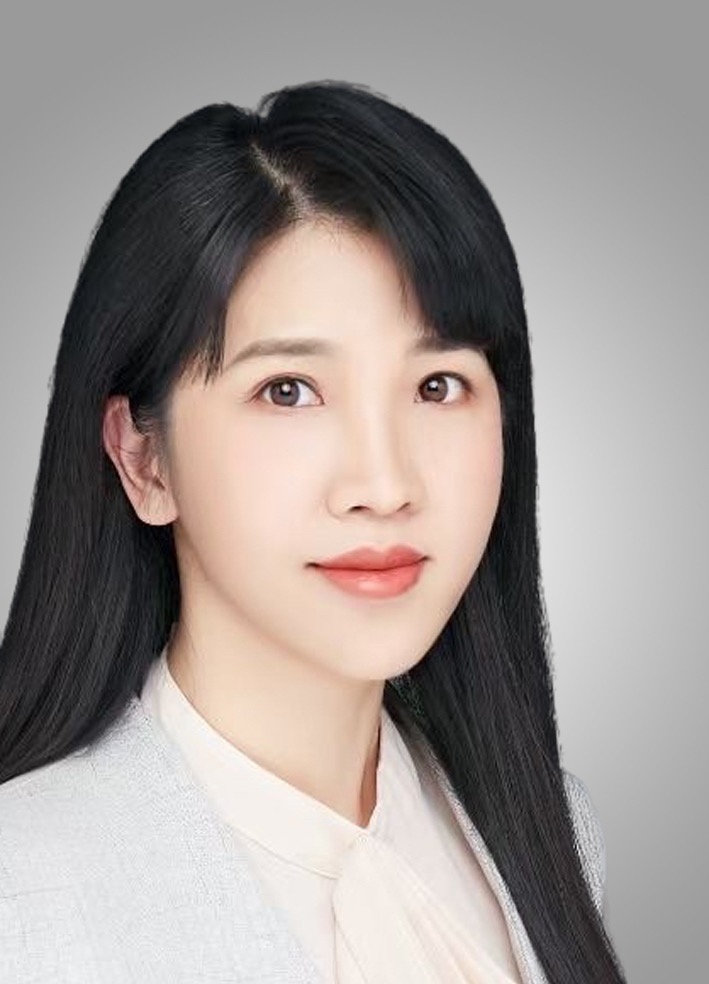}}]{Huijia Zhu} graduated with a master's degree from the Department of Computer Science and Technology at Harbin Institute of Technology in 2007. Her research focus is on Natural Language Understanding. In her professional career, she currently works at Ant Group, specializing in algorithm research related to content security. Her primary research areas include Natural Language Processing, Multimodal Understanding, and the practical application of large models.\end{IEEEbiography}

\vspace{-2.5cm}

\begin{IEEEbiography}[{\includegraphics[width=1in,height=1.25in,clip,keepaspectratio]{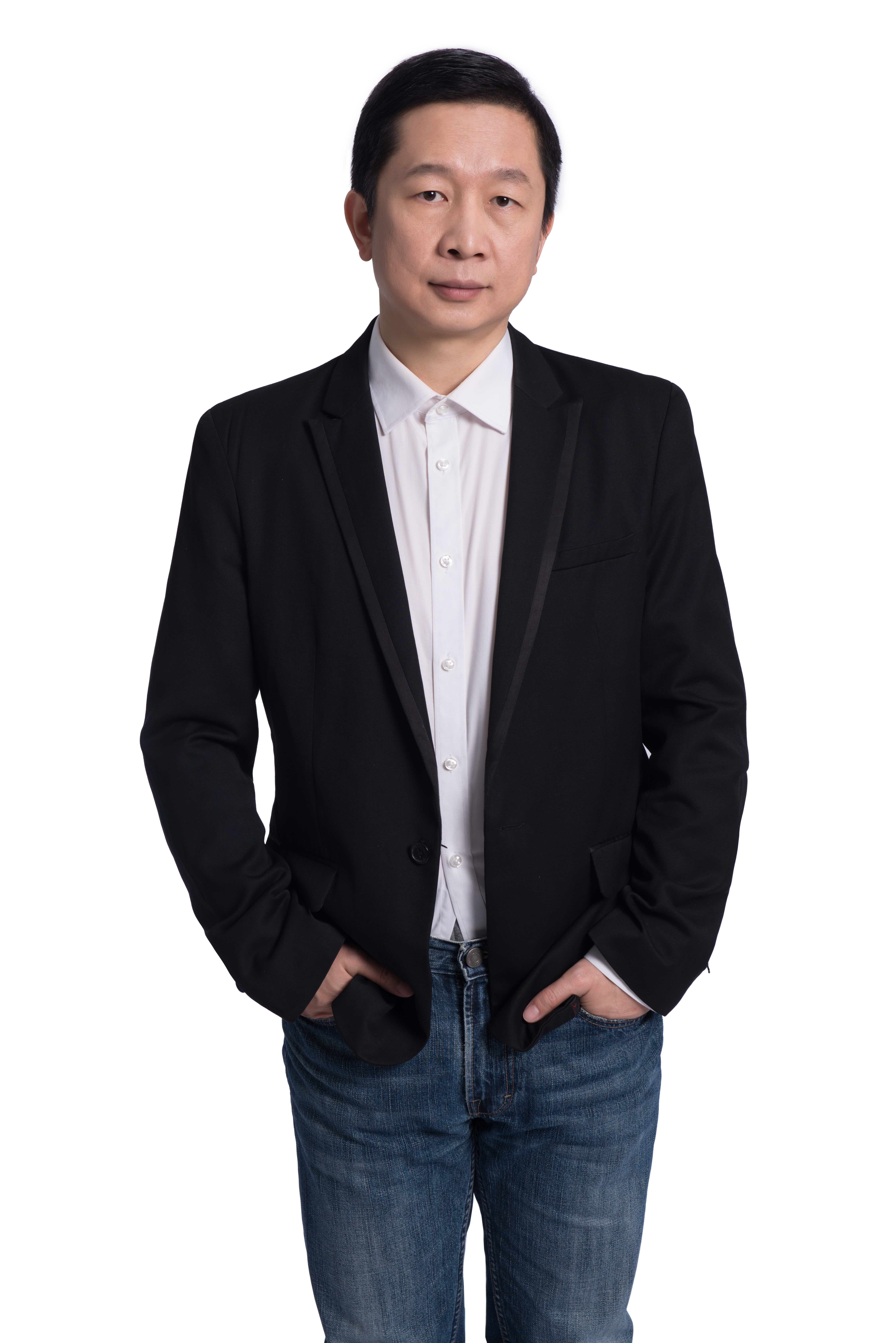}}]{Weiqiang Wang}, Chief Scientist of the Ant Group Security Laboratory. He holds a Master's degree in Computer Science and a Ph.D. in Materials Science from the University of Southern California, USA. His current main research focus is on the application of machine learning AI algorithms in digital financial risk control and security. He is responsible for the research and application of AI algorithms in risk control and security across various Ant Group businesses. During his tenure, he led the team to design and establish a risk control model system for fraud, abuse, and cheating, creating the fifth-generation intelligent risk control decision engine for Ant Group. He has published multiple research results in top international journals and conferences and has obtained several national and international patents. \end{IEEEbiography}

\vspace{-2.5cm}

\begin{IEEEbiography}[{\includegraphics[width=1in,height=1.25in,clip,keepaspectratio]{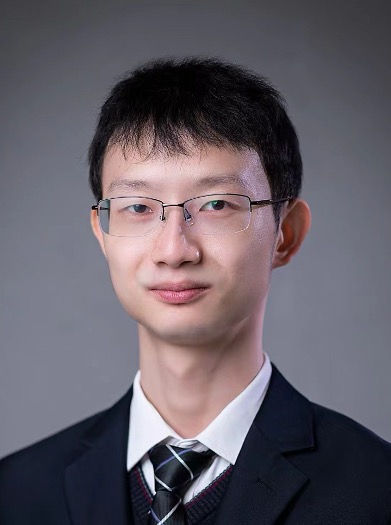}}]{Zhuosheng Zhang} received his Bachelor's degree from Wuhan University in 2016, and his M.S. and Ph.D. degrees from Shanghai Jiao Tong University in 2020 and 2023, respectively. He is currently a tenure-track assistant professor at Shanghai Jiao Tong University. He was a research intern at Amazon AWS, Microsoft Research, Langboat Technology, NICT (Japan), and IBM. His research interests include natural language processing, large language models, and language agents. 
He has published research papers in leading journals and conferences, such as TPAMI, TNNLS, TASLP, ICLR, ICML, ACL, AAAI, EMNLP, and COLING. 
He was the recipient of the WAIC 2024 Youth Outstanding Paper Award, WAIC 2024 YunFan Award, and the Global Top 100 Chinese Rising Stars in Artificial Intelligence. He serves as an action editor for ACL Rolling Review and standing reviewer for TACL. He served as a (senior) area chair for conferences such as NeurIPS, ACL, IJCAI, EMNLP, and COLING.\end{IEEEbiography}

\vspace{-2.5cm}

\begin{IEEEbiography}[{\includegraphics[width=1in,height=1.25in,clip,keepaspectratio]{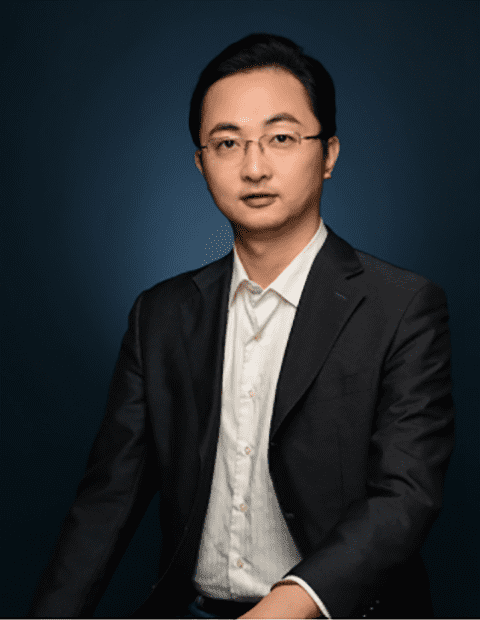}}]{Shilin Wang} (Senior Member, IEEE) received the B.Eng. degree in electrical and electronic engineering from Shanghai Jiao Tong University, Shanghai, China, in 2001, and the Ph.D. degree from the Department of Computer Engineering and Information Technology, City University of Hong Kong, in 2004. Since 2004, he has been with the School of Electric Information and Electronic Engineering, Shanghai Jiao Tong University, where he is currently a Professor. His research interests include image processing and pattern recognition.\end{IEEEbiography}